\documentclass[10pt,twocolumn,letterpaper]{article}

\usepackage{iccv}
\usepackage{times}
\usepackage{epsfig}
\usepackage{graphicx}
\usepackage{amsmath}
\usepackage{amssymb}
\DeclareMathOperator{\sign}{sgn}
\DeclareMathOperator*{\argmin}{arg\,min}

\usepackage{fancyhdr}

\usepackage{umoline}
\newcommand{\todo}[2]{#1}

\usepackage[pagebackref=true,breaklinks=true,letterpaper=true,colorlinks,bookmarks=false]{hyperref}

\iccvfinalcopy 

\def\iccvPaperID{1046} 

\begin{document}

\title{Universal Adversarial Perturbations Against Semantic Image Segmentation}

\author{Jan Hendrik Metzen\\
Bosch Center for Artificial Intelligence, Robert Bosch GmbH \\
{\tt\small janhendrik.metzen@de.bosch.com}
\and
Mummadi Chaithanya Kumar\\
University of Freiburg\\
{\tt\small chaithu0536@gmail.com}
\and
Thomas Brox\\
University of Freiburg\\
{\tt\small brox@cs.uni-freiburg.de}
\and
Volker Fischer\\
Bosch Center for Artificial Intelligence, Robert Bosch GmbH\\
{\tt\small volker.fischer@de.bosch.com}
}

\maketitle
\thispagestyle{fancy}
\fancyhf{}
\renewcommand{\headrulewidth}{0.0pt}
\lfoot{\tiny \copyright 2017 IEEE. Personal use of this material is permitted. Permission from IEEE must be obtained for all other uses, in any current or future media, including reprinting/republishing this material for advertising or promotional purposes, creating new collective works, for resale or redistribution to servers or lists, or reuse of any copyrighted component of this work in other works.} 

\begin{abstract}
While deep learning is remarkably successful on perceptual tasks, it was also shown to be vulnerable to adversarial perturbations of the input. These perturbations denote noise added to the input that was generated specifically to fool the system while being quasi-imperceptible for humans. More severely, there even exist universal perturbations that are input-agnostic but fool the network on the majority of inputs. While recent work has focused on image classification, this work proposes attacks against semantic image segmentation: we present an approach for generating (universal) adversarial perturbations that make the network yield a desired target segmentation as output. We show empirically that there exist barely perceptible universal noise patterns which result in nearly the same predicted segmentation for arbitrary inputs. Furthermore, we also show the existence of universal noise which removes a target class (e.g., all pedestrians) from the segmentation while leaving the segmentation mostly unchanged otherwise.
\end{abstract}

\section{Introduction} \label{Section:Introduction}

While deep learning has led to significant performance increases for numerous visual perceptual tasks \cite{he_deep_2015,long_fully_2015,ren_faster_2015,taigman_deepface:_2014} and is relatively robust to random noise \cite{fawzi_robustness_2016}, several studies have found it to be vulnerable to adversarial perturbations \cite{szegedy_intriguing_2013,goodfellow_explaining_2015,Moosavi-Dezfooli_2016_CVPR,sharif_accessorize_2016,carlini_towards_2016}. Adversarial attacks involve generating slightly perturbed versions of the input data that fool the classifier (i.e., change its output) but stay almost imperceptible to the human eye. Adversarial perturbations transfer between different network architectures, and networks trained on disjoint subsets of data \cite{szegedy_intriguing_2013}. Furthermore, Papernot et al.\,\cite{papernot_practical_2016} showed that adversarial examples for a network of unknown architecture can be constructed by training an auxiliary network on similar data and exploiting the transferability of adversarial examples. 

\begin{figure}[t]
\begin{center}
\setlength{\tabcolsep}{3pt}
\begin{tabular}{cc}
(a) Image & (b) Prediction \\
\includegraphics[width=.5\linewidth]{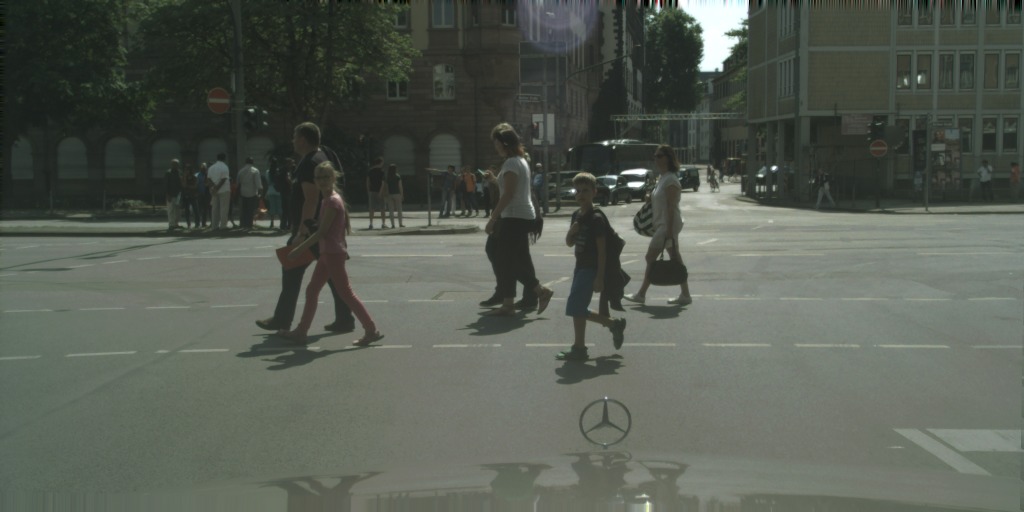} & 
\includegraphics[width=.5\linewidth]{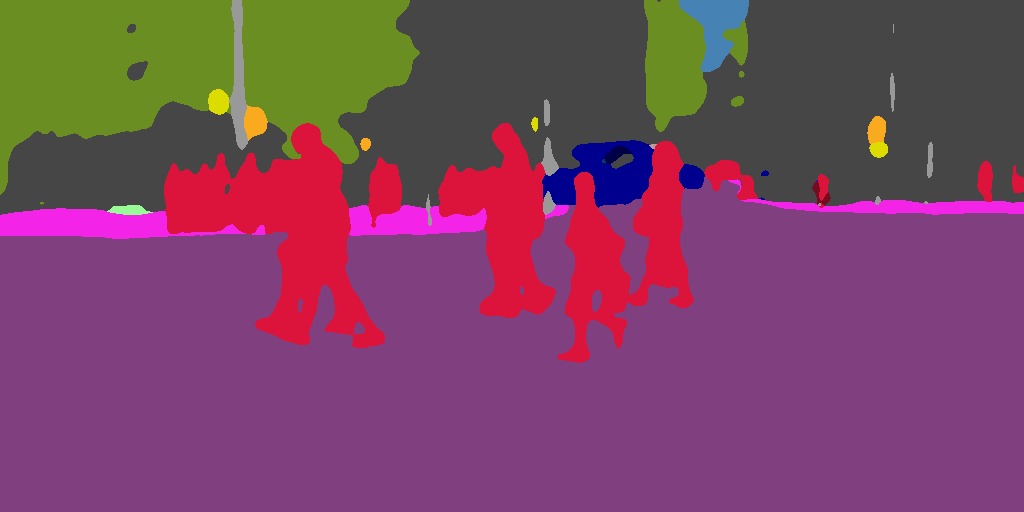}\\
(c) Adversarial Example & (d) Prediction \\
\includegraphics[width=.5\linewidth]{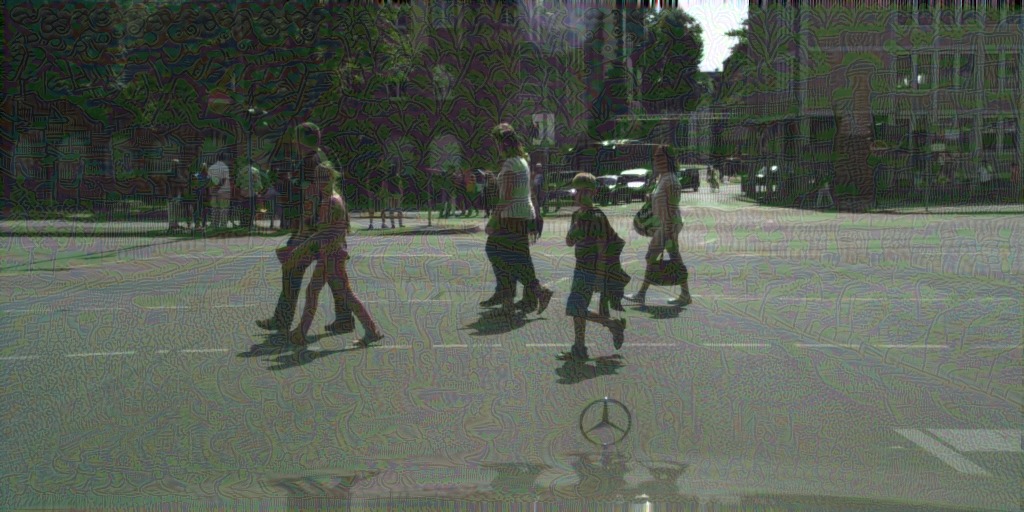} & 
\includegraphics[width=.5\linewidth]{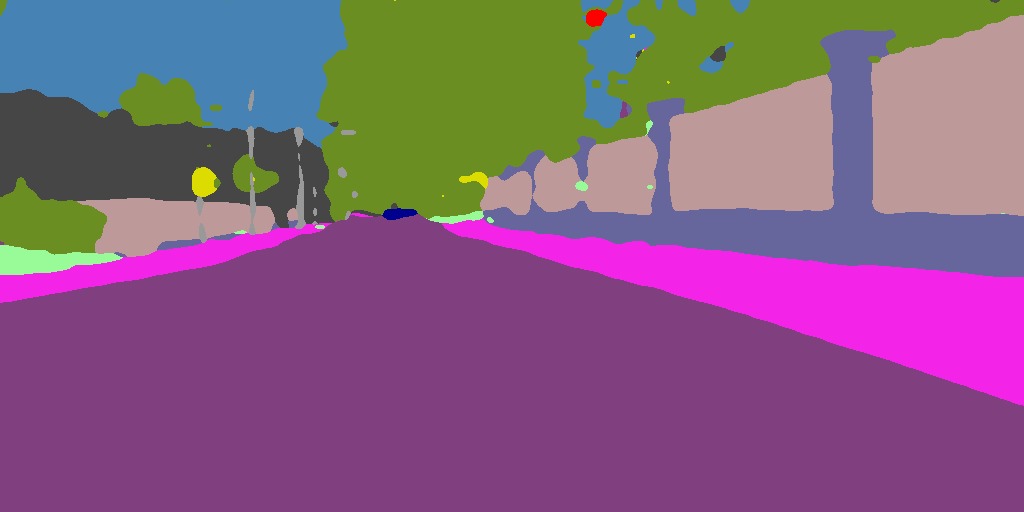} \\
\end{tabular}
\end{center}
\caption{The upper row shows an image from the validation set of Cityscapes and its prediction. The lower row shows the image perturbed with universal adversarial noise and the resulting prediction. Note that the prediction would look very similar for other images when perturbed with the same noise (see Figure \ref{fig:illustration_static_targets}).}
\label{fig:motivating_illustration}
\end{figure}

Prior work on adversarial examples focuses on the task of image classification. In this paper, we investigate the effect of adversarial attacks on tasks involving a localization component, more specifically: semantic image segmentation. Semantic image segmentation is an important methodology for scene understanding that can be used for example for automated driving, video surveillance, or robotics. With the wide-spread applicability in those domains comes the risk of being confronted with an adversary trying to fool the system. Thus, studying adversarial attacks on semantic segmentation systems deployed in the physical world becomes an important problem.

Adversarial attacks that aim at systems grounded in the physical world should be physically realizable and inconspicuous \cite{sharif_accessorize_2016}. One prerequisite for \emph{physical realizability} is that perturbations do not depend on the specific input since this input is not known in advance when the perturbations (which need to be placed in the physical world) are determined. This work proposes a method for generating image-agnostic \emph{universal perturbations}. Universal perturbations have been proposed by Moosavi-Dezfooli et al.\,\cite{moosavi-dezfooli_universal_2017}; however, we extend the idea to the task of semantic image segmentation. We leave further prerequisites for physical realizability as detailed by Sharif et al.\,\cite{sharif_accessorize_2016} to future work.

An attack is \emph{inconspicuous} if it does not raise the suspicion of humans monitoring the system (at least not under cursory investigation). This requires that the system inputs are modified only subtly, and, for a semantic image segmentation task, also requires that system output (the scene segmentation) looks mostly as a human would expect for the given scene. If an adversary's objective is to remove all occurrences of a specific class (e.g., an adversary trying to hide all pedestrians to deceive an emergency braking system) then the attack is maximally inconspicuous if it leaves the prediction for all other classes unchanged and only hides the target class. We present one adversarial attack which explicitly targets this \emph{dynamic target segmentation} scenario.

While inconspicuous attacks require that target scenes mostly match what a human expects, we also present an attack yielding an \emph{static target segmentations}. This attack generates universal perturbations that let the system output always essentially the same segmentation regardless of the input, even when the input is from a completely different scene (see Figure \ref{fig:motivating_illustration}). The main motivation for this experiment is to show how fragile current approaches for semantic segmentation are when confronted with an adversary. In practice, such attacks could be used in scenarios in which a static camera monitors a scene (for instance in surveillance scenarios) as it would allow an attacker to always output the segmentation of the background scene and blend out all activity like, e.g., burglars robbing a jewelry shop.

We summarize our main contributions as follows:
\begin{itemize}
\setlength{\itemsep}{0pt}
\setlength{\parsep}{0pt}
\item We show the existence of (targeted) universal perturbations for semantic image segmentation models. Their existence was not clear a priori because the receptive fields of different output elements largely overlap. Thus perturbations cannot be chosen independently for each output target. This makes the space of adversarial perturbations for semantic image segmentation presumably smaller than for recognition tasks like image classification and the existence of universal perturbations even more surprising.
\item We propose two efficient methods for generating these universal perturbations. These methods optimize the perturbations on a training set. The objective of the first methods is to let the network yield a fixed target segmentation as output. The second method's objective is to leave the segmentation unchanged except for removing a designated target class.
\item We show empirically that the generated perturbations are generalizable: they fool unseen validation images with high probability. Controlling the capacity of universal perturbations is important for achieving this generalization from small training sets.
\item We show that universal perturbations generated for a fixed target segmentation have a local structure that resembles the target scene (see Figure \ref{fig:universal_static_perturbation}).
\end{itemize}

\section{Background}

Let $f_\theta$ be a function with parameters $\theta$. Moreover, let $\mathbf{x}$ be an input of $f_\theta$, $f_\theta(\mathbf{x})$ be the output of $f_\theta$, and $\mathbf{y}^\text{true}$ be the corresponding ground-truth target. More specifically for the scenario studied in this work, $f_\theta$ denotes a deep neural network, $\mathbf{x}$ an image, $f_\theta(\mathbf{x})$ the conditional probability $p(\mathbf{y} \vert \mathbf{x}; \theta)$ encoded as a class probability vector, and $\mathbf{y}^\text{true}$ a one-hot encoding of the class. Furthermore, let
$J_\text{cls}(f_\theta(\mathbf{x}), \mathbf{y}^\text{true})$ be the basic classification loss such as cross-entropy. We assume that $J_\text{cls}$ is differentiable with respect to $\theta$ and with respect to $\mathbf{x}$.

\subsection{Semantic Image Segmentation} \label{Section:BackgroundSemSeg}

Semantic image segmentation denotes a dense prediction task that addresses the ``what is where in an image?'' question by assigning a class label to each pixel of the image. Recently, deep learning based approaches (oftentimes combined with conditional random fields) have become the dominant and best performing class of methods for this task \cite{long_fully_2015,liu_parsing_2015, zheng_crf_rnn_2015,chen_crf_2015,fisher_dilated_2016,chen_attention_2016}. In this work, we focus on one of the first and most prominent architectures, the fully convolutional network architecture FCN-8s introduced by Long et al.\,\cite{long_fully_2015} for the VGG16 model \cite{simonyan_vgg_2015}. 

The FCN-8s architecture can roughly be divided into two parts: an encoder part which transforms a given image into a low resolution semantic representation and a decoder part which increases the localization accuracy and yields the final semantic segmentation at the  resolution of the input image. The encoder part is based on a VGG16 pretrained on ImageNet \cite{russakovsky_imagenet_2015} where the fully connected layers are reinterpreted as convolutions making the network ``fully convolutional''. The output of the last encoder layer can be interpreted as a low-resolution semantic representation of the image and is the input to five upsampling layers which recover the high spatial resolution of the image via successive bilinear-interpolation (FCN-32s). For FCN-8s, additionally two parallel paths merge higher-resolution, less abstract layers of the VGG16 into the upsampling path via convolutions and element-wise summation. This enables the network to utilize features with a higher spatial resolution.


\subsection{Adversarial Examples} \label{Section:Background_AdvExamples}

Let $\mathbf{\xi}$ denote an \emph{adversarial perturbation} for an input $\mathbf{x}$ and let $\mathbf{x}^{\text{adv}} = \mathbf{x} + \mathbf{\xi}$ denote the corresponding \emph{adversarial example}. The objective of an adversary is to find a perturbation $\mathbf{\xi}$ which changes the output of the model in a desired way. For instance the perturbation can either make the true class less likely or a designated target class more likely. At the same time, the adversary typically tries to keep $\mathbf{\xi}$ quasi-imperceptible by, e.g., bounding its $\ell_\infty$-norm.

The first method for generating adversarial examples was proposed by Szegedy et al.\,\cite{szegedy_intriguing_2013}. While this method was able to generate adversarial examples successfully for many  inputs and networks, it was also relatively slow computationally since it involved an L-BFGS-based optimization. Since then, several methods for generating adversarial examples have been proposed. These methods either maximize the predicted probability for all but the true class or minimize the probability of the true class. 

Goodfellow et al.\,\cite{goodfellow_explaining_2015} proposed a non-iterative and hence fast method for computing adversarial perturbations. This \emph{fast gradient-sign method} (FGSM) defines an adversarial perturbation as the direction
in image space which yields the highest increase of the linearized cost function under $\ell_{\infty}$-norm. 
This can be achieved by performing one step in the gradient sign's direction with step-width $\varepsilon$:
$$ \mathbf{\xi} = \varepsilon\sign(\nabla_{\mathbf{x}}J_\text{cls}(f_\theta(\mathbf{x}), \mathbf{y}^\text{true})) $$

Here, $\varepsilon$ is a hyper-parameter governing the distance between original image and adversarial image. FGSM is a targeted method. This means that the adversary is solely trying to make the predicted probability of the true class smaller. However, it does not control which of the other classes becomes more probable. 

Kurakin et al.\,\cite{kurakin_adversarial_2016} proposed an extension of FGSM which is iterative and targeted. The proposed \emph{least-likely method} (LLM) makes the least likely class $\mathbf{y}^\text{LL} = \arg\min_\mathbf{y} p(\mathbf{y} \vert \mathbf{x})$ under the prediction of the model more probable.  LLM is in principle not specific for the least-likely class $\mathbf{y}^{LL}$; it can rather be used with an arbitrary target class $\mathbf{y}^\text{target}$. The method tries to find $\mathbf{x}^{\text{adv}}$ which maximizes the predictive probability of class $y^\text{target}$ under $f_\theta$. This can be achieved by the following iterative procedure:
\begin{small}
\begin{align*}
\mathbf{\xi}^{(0)} &= \mathbf{0}, \\
\mathbf{\xi}^{(n+1)} &= \text{Clip}_{\varepsilon}\left\{\mathbf{\xi}^{(n)} - \alpha\sign(\nabla_{\mathbf{x}}J_\text{cls}(f_\theta(\mathbf{x} + \mathbf{\xi}^{(n)}),\mathbf{y}^\text{target}))\right\}
\end{align*}
\end{small}

Here $\alpha$ denotes a step size and all entries of $\mathbf{\xi}$ are clipped after each iteration such that their absolute value remains smaller than $\varepsilon$. We use $\alpha=1$ throughout all experiments. Concurrent with this work, adversarial examples have been extended to semantic image segmentation and object detection \cite{xie_adversarial_2017,fischer_adversarial_2017}. Moreover, training with adversarial examples has been applied to mammographic mass segmentation to reduce overfitting \cite{zhu_adversarial_2016}.

For the methods outlined above, the adversarial perturbation $\mathbf{\xi}$ depends on the input $\mathbf{x}$. 
Recently, Moosavi-Dezfooli et al.\,\cite{moosavi-dezfooli_universal_2017} proposed a method for generating \emph{universal, image-agnostic perturbations} $\mathbf{\Xi}$ that, when added to arbitrary data points, fool deep nets on a large fraction of images. The method for generating these adversarial perturbations is based on the adversarial attack method DeepFool \cite{Moosavi-Dezfooli_2016_CVPR}. DeepFool is applied to a set of $m$ images (the train set). These images are presented sequentially in a round-robin manner to DeepFool. For the first image, DeepFool identifies a standard image-dependent perturbation. For subsequent images, it is checked whether adding the previous adversarial perturbation already fools the classifier; if yes the algorithm continues with the next image, otherwise it updates the perturbation using DeepFool such that also the current image becomes adversarial. The algorithm stops once the perturbation is adversarial on a large fraction of the train set. 

The authors show impressive results on ImageNet \cite{russakovsky_imagenet_2015}, where they show that the perturbations are adversarial for a large fraction of test images, which the method did not see while generating the perturbation.  One potential short-coming of the approach is that the attack is not targeted, i.e., the adversary cannot control which class the classifier shall assign to an adversarial example. Moreover, for high-resolution images and a small train set, the perturbation might overfit the train set and not generalize to unseen test data since the number of ``tunable parameters'' is proportional to the number of pixels. Thus, high-resolution images will need a large train set and a large computational budget. In this paper, we propose a method which overcomes these shortcomings.

\section{Adversarial Perturbations Against Semantic Image Segmentation}

For semantic image segmentation, the loss is a sum over the spatial dimensions $(i, j) \in \mathcal{I}$ of the target such as:
$$J_{\text{ss}}(f_\theta(\mathbf{x}), \mathbf{y}) = 
\frac{1}{\vert \mathcal{I} \vert} \sum\limits_{(i,j) \in \mathcal{I}} J_\text{cls}(f_\theta(\mathbf{x})_{ij}, \mathbf{y}_{ij}).$$
In this section, we describe how to find an input $\mathbf{x}^{\text{adv}}$ for $f_\theta$ such that $J_{\text{ss}}(f_\theta(\mathbf{x}^{\text{adv}}), \mathbf{y}^{\text{target}})$ becomes minimal, i.e., how an adversary can do quasi-imperceptible changes to the input such that it achieves a desired target segmentation $\mathbf{y}^{\text{target}}$. We start by describing how an adversary can choose $\mathbf{y}^{\text{target}}$.

\subsection{Adversarial Target Generation} \label{Subsection:AdversarialTargetGeneration}
In principle, an adversary may choose $\mathbf{y}^{\text{target}}$ arbitrarily. Crucially, however, an adversary may not choose $\mathbf{y}^{\text{target}}$ based on $\mathbf{y}^{\text{true}}$ since the ground-truth is also unknown to the adversary. Instead, the adversary may use $\mathbf{y}^{\text{pred}} = f_\theta(\mathbf{x})$ as basis as we assume that the adversary has access to $f_\theta$.

As motivated in Section \ref{Section:Introduction}, typical scenarios involve an adversary whose primary objective is to hide certain kinds of objects such as, e.g., pedestrians. As a secondary objective, an adversary may try to perform attacks that are \emph{inconspicuous}, i.e., do not call the attention of humans monitoring the system (at least not under cursory investigation) \cite{sharif_accessorize_2016}. Thus the input must be modified only subtly. For a semantic image segmentation task, however, it is also required that the output of the system looks mostly as a human would expect for the given scene. This can be achieved, for instance, by keeping $\mathbf{y}^{\text{target}}$ as similar as possible to $\mathbf{y}^{\text{pred}}$ where the primary objective does not apply. We define two different ways of generating the target segmentation:

\begin{figure*}[t]
\centering
\includegraphics[width=\linewidth]{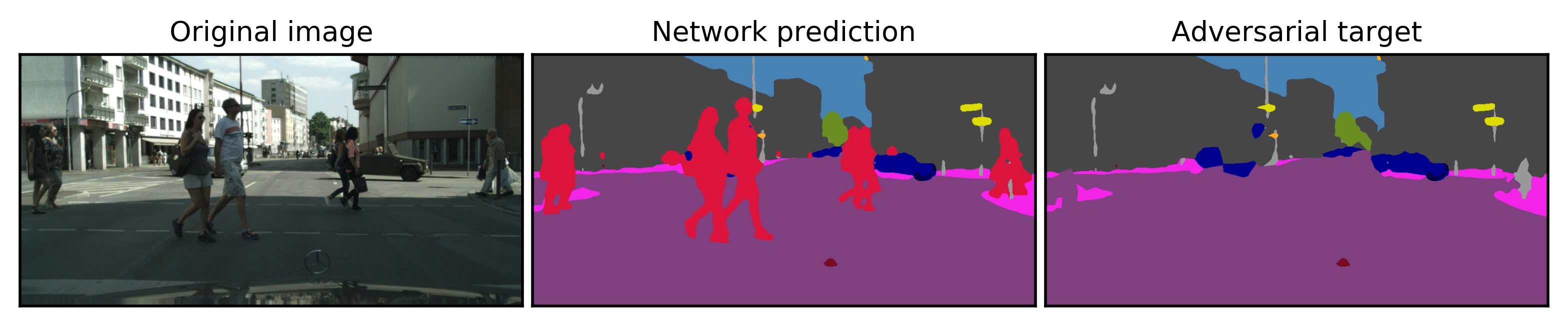}
\caption{Illustration of an adversary generating a dynamic target segmentation for hiding pedestrians.}
\label{fig:adversarial_target}
\end{figure*}

\paragraph{Static target segmentation:} In this scenario, the adversary defines a fixed segmentation, such as
the system's prediction at a time step $t_0$, as target for all subsequent time steps: $\mathbf{y}^{\text{target}}_{t} = \mathbf{y}^{\text{pred}}_{t_0} \,\forall t > t_0$. This target segmentation is suited for instance in situations where an adversary wants to attack a system based on a static camera and wants to hide suspicious activity in a certain time span $t > t_0$ that had not yet started at time $t_0$.

\paragraph{Dynamic target segmentation:}
In situations involving ego-motion, a static target segmentation is not suited as it would not account for changes in the scene caused by the movement of the camera. In contrast, \emph{dynamic target segmentation} aims at keeping the network's segmentation unchanged with the exception of removing certain target classes. Let $o$ be the class of objects the adversary wants to hide, and let $\mathcal{I}_o = \left\{ (i, j) \mid f_\theta(x_{ij}) = o \right\}$ and $\mathcal{I}_{bg} = \mathcal{I} \setminus \mathcal{I}_o$. We assign $\mathbf{y}^{\text{target}}_{ij} = \mathbf{y}^{\text{pred}}_{ij}$ for all $(i,j) \in \mathcal{I}_{bg}$, and $\mathbf{y}^{\text{target}}_{ij} = \mathbf{y}^{\text{pred}}_{i'j'}$ for all $(i,j) \in \mathcal{I}_{o}$ with $i', j' = \argmin\limits_{i', j' \in \mathcal{I}_{bg}} (i' - i)^2 + (j' - j)^2$. The latter corresponds to filling the gaps left in the target segmentation by removing elements predicted to be $o$ using a nearest-neighbor heuristic. An illustration of the adversarial target generation is shown in Figure \ref{fig:adversarial_target}.

\subsection{Image-Dependent Perturbations} \label{Subsection:Adversarial_Attack}
Before turning to image-agnostic universal perturbations, we first define how an adversary might choose an image-dependent perturbation. Given $\mathbf{y}^{\text{target}}$, we formulate the objective of the adversary as follows:
$$\mathbf{\xi}_{\text{adv}} = \arg\min_{\mathbf{\xi}'} J_{\text{ss}}(f_\theta(\mathbf{x} + \mathbf{\xi}'), \mathbf{y}^\text{target}) 
\; \text{s.t.} \; \vert \mathbf{\xi}_{ij}' \vert \leq \varepsilon$$

The constraint limits the adversarial example $\mathbf{x} + \mathbf{\xi}'$ to have at most an  $\ell_\infty$-distance of $\varepsilon$ to $\mathbf{x}$. Let $\text{Clip}_{\mathbf{\varepsilon}}\left\{\mathbf{\xi}\right\}$ implement the constraint $\vert \mathbf{\xi}_{ij} \vert \leq \varepsilon$ by clipping all entries of $\xi$ to have at most an absolute value of $\varepsilon$. Based on this, we can define a targeted iterative adversary analogously to the least-likely method (see Section \ref{Section:Background_AdvExamples}):

\begin{small}
\begin{align*}
\mathbf{\xi}^{(0)} &= \mathbf{0},\\
\mathbf{\xi}^{(n+1)} &= \text{Clip}_{\mathbf{\varepsilon}}\left\{\mathbf{\xi}^{(n)} - \alpha\sign(\nabla_{x}J_{\text{ss}}(f_\theta(\mathbf{x} + \mathbf{\xi}^{(n)}),\mathbf{y}^\text{target}))\right\}
\end{align*}
\end{small}

An alternative formulation which takes into consideration that the primary objective (hiding objects) and the secondary objective (being inconspicuous) are not necessarily equally important can be achieved by a modified version of the loss including a weighting parameter $\omega$:

\begin{small}
\begin{align*}
J_{\text{ss}}^\omega(f_\theta(\mathbf{x}), \mathbf{y}^\text{target}) = 
\frac{1}{\vert \mathcal{I} \vert} \{&\omega \sum\limits_{(i,j) \in \mathcal{I}_o} J_\text{cls}(f_\theta(\mathbf{x})_{ij}, \mathbf{y}_{ij}^\text{target}) + \\
&(1 - \omega) \sum\limits_{(i,j) \in \mathcal{I}_{bg}} J_\text{cls}(f_\theta(\mathbf{x})_{ij}, \mathbf{y}_{ij}^\text{target})\}
\end{align*}
\end{small}

Here, $\omega=1$ lets the adversary solely focus on removing target-class predictions, $\omega=0$ forces the adversary only to keep the background constant, and $J_{\text{ss}}^\omega = 0.5 J_{\text{ss}}$ for $\omega=0.5$.

An additional issue for $J_{\text{ss}}$ (and $J_{\text{ss}}^\omega$) is that there is potentially competition between different target pixels, i.e., the gradient of the loss for $(i_1, j_1)$ might point in the opposite direction as the loss gradient for $(i_2, j_2)$. Standard classification losses such as the cross entropy in general encourage target predictions which are already correct to become more confident as this reduces the loss. This is not necessarily desirable in face of competition between different targets. The reason for this is that loss gradients for making correct predictions more confident might counteract loss gradients which would make wrong predictions correct. Note that this issue does not exist for adversaries targeted at image classification as there is essentially only a single target output. To address this issue, we set the loss of target pixels which are predicted as the desired target  with a confidence above $\tau$ to $0$ \cite{wu_high-performance_2016}. Throughout this paper, we use $\tau=0.75$.

\subsection{Universal Perturbations}
In this section, we propose a method for generating \emph{universal} adversarial perturbations $\mathbf{\Xi}$ in the context of semantic segmentation. The general setting is that we generate $\mathbf{\Xi}$  on a set of $m$ training inputs $\mathcal{D}^\text{train}=\{(\mathbf{x}^{(k)}, \mathbf{y}^{\text{target},k})\}_{k=1}^m$, where $\mathbf{y}^{\text{target},k}$ was generated with either of the two methods presented in Section \ref{Subsection:AdversarialTargetGeneration}. We are interested in the generalization of $\mathbf{\Xi}$ to test inputs $\mathbf{x}$ for which it was not optimized and for which no target $\mathbf{y}^\text{target}$ exists. This generalization to inputs for which no target exists is required because 
generating $\mathbf{y}^\text{target}$ would require evaluating $f_\theta$ which might not be possible at test time or under real-time constraints. We propose the following extension of the attack presented in Section \ref{Subsection:Adversarial_Attack}:

\begin{align*}
\mathbf{\Xi}^{(0)} &= \mathbf{0},\\
\mathbf{\Xi}^{(n+1)} &= \text{Clip}_{\mathbf{\varepsilon}}\left\{\mathbf{\Xi}^{(n)} - \alpha\sign(\nabla^{\mathcal{D}}(\mathbf{\Xi}))\right\},
\end{align*}
with $\nabla^{\mathcal{D}}(\mathbf{\Xi}) = \frac{1}{m} \sum\limits_{k=1}^m\nabla_{x}J_{\text{ss}}^\omega(f_\theta(\mathbf{x}^{(k)} + \mathbf{\Xi}),\mathbf{y}^{\text{target},k})$ being the loss gradient averaged over the entire training data. A potential issue of this approach is overfitting to the training data which would reduce generalization of $\mathbf{\Xi}$ to unseen inputs. Overfitting is actually likely given that $\mathbf{\Xi}$ has the same dimensionality as the input image and is thus high-dimensional. We adopt a relatively simple regularization approach by enforcing $\mathbf{\Xi}$ to be periodic in both spatial dimensions. More specifically, we enforce for all $i,j \in \mathcal{I}$ the constraints $\mathbf{\Xi}_{i,j} = \mathbf{\Xi}_{i+h,j}$ and $\mathbf{\Xi}_{i,j} = \mathbf{\Xi}_{i,j+w}$ for a predefined spatial periodicity $h, w$. This can be achieved by optimizing a proto-perturbation $\mathbf{\hat{\Xi}}$ of size $h \times w$ and tile it to the full $\mathbf{\Xi}$. This results in a gradient averaged over the training data and all tiles: 
$$\nabla^{\mathcal{D}}(\mathbf{\hat{\Xi}}) = \frac{1}{mRS} \sum\limits_{r=1}^R\sum\limits_{s=1}^S\sum\limits_{k=1}^m\nabla_{x}J_{\text{ss}}^\omega(f_\theta(\mathbf{x}^{(k)}_{[r,s]} + \mathbf{\hat{\Xi}}),\mathbf{y}^{\text{target},k}_{[r,s]}),$$
with $R$, $S$ denoting the number of tiles per dimension and $[r, s] = \{i, j \mid [rh \leq i < (r+1)h] \wedge [sw \leq j < (s+1)w]\}$.

As we will show in Section \ref{Section:ExperimentalResults}, the quality of the generated universal perturbation depends crucially on the size $m$ of the train set. As our method for generating universal perturbations does not require ground-truth labels, we may in principle use arbitrary large unlabeled data sets. Nevertheless, we also investigate how well universal perturbations can be generated for small $m$ since large $m$ requires considerable computational resources and also more queries to $f_\theta$, which might increase monetary costs or the risk of being identified.

\section{Experimental Results} \label{Section:ExperimentalResults}

We evaluated the proposed adversarial attacks against semantic image segmentation on the Cityscapes dataset \cite{Cordts2016Cityscapes}, which consists of $3475$ publicly available labeled RGB images ($2975$ for training and $500$ for validation) with a resolution of $2048\times1024$ pixels from $44$ different cities. We used the pixel-wise fine annotations covering $19$ frequent classes. For computational reasons, all images and labels were downsampled to a resolution of $1024\times512$ pixels, where for images a bilinear interpolation and for labels a nearest-neighbor approach was used for down-sampling. We trained the FCN-8s network architecture (see Section \ref{Section:BackgroundSemSeg}) for semantic image segmentation on the whole training data and achieved a class-wise intersection-over-union (IoU) on the validation data of 64.8\%.

We generated the universal perturbations on (subsets of) the training data and evaluated them on unseen validation data. When not noted otherwise, we used $\varepsilon=10$ in the experiments. This value of $\varepsilon$ was also used by Moosavi-Dezfooli et al.\,\cite{moosavi-dezfooli_universal_2017} and corresponds to a level of noise which is only perceptible for humans at closer inspection. Moreover, we set the number of iterations to $n=60$.

\begin{figure*}
\begin{center}
\setlength{\tabcolsep}{3pt}
\begin{tabular}{cccc}
(a) image 1 & (b)  pred. image 1 & (c) image 2 & (d) pred. image 2 \\
\includegraphics[width=.23\linewidth]{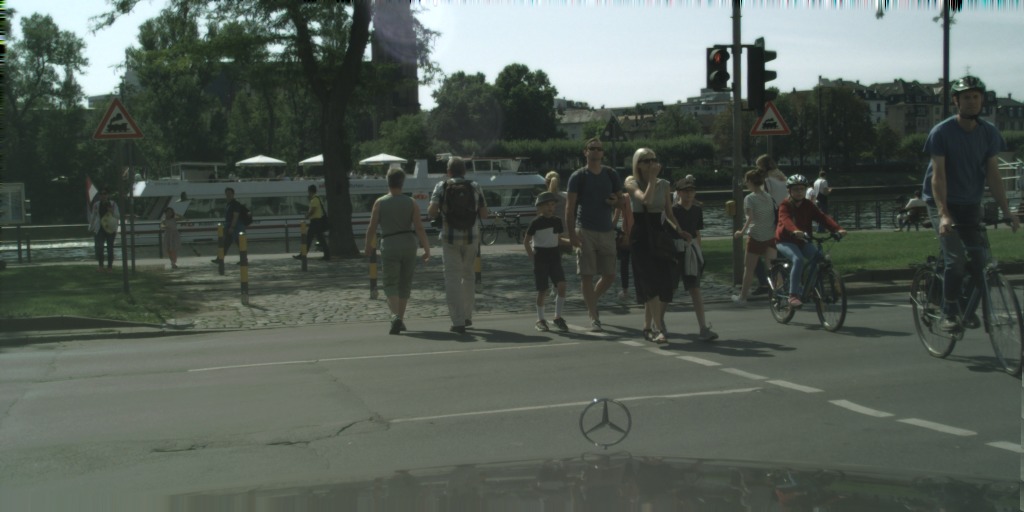} & 
\includegraphics[width=.23\linewidth]{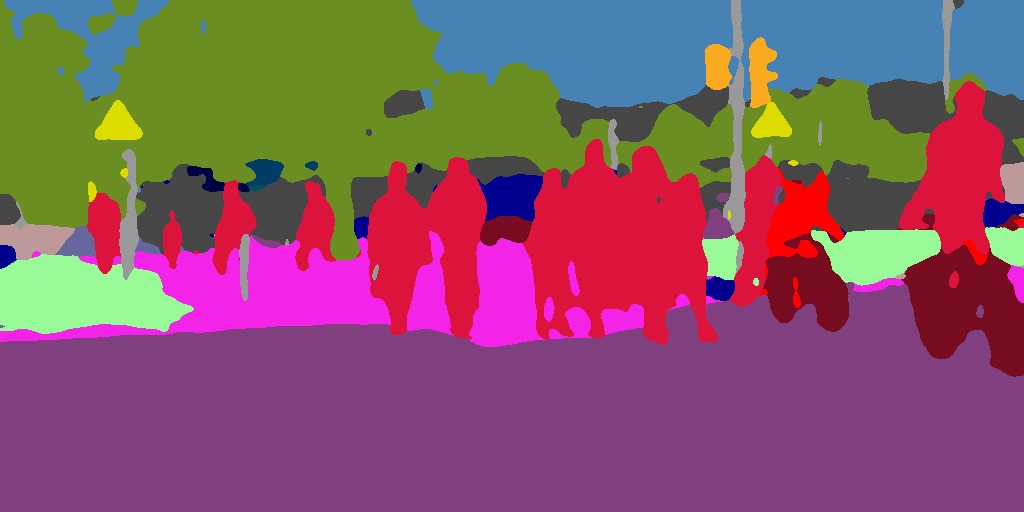} & 
\includegraphics[width=.23\linewidth]{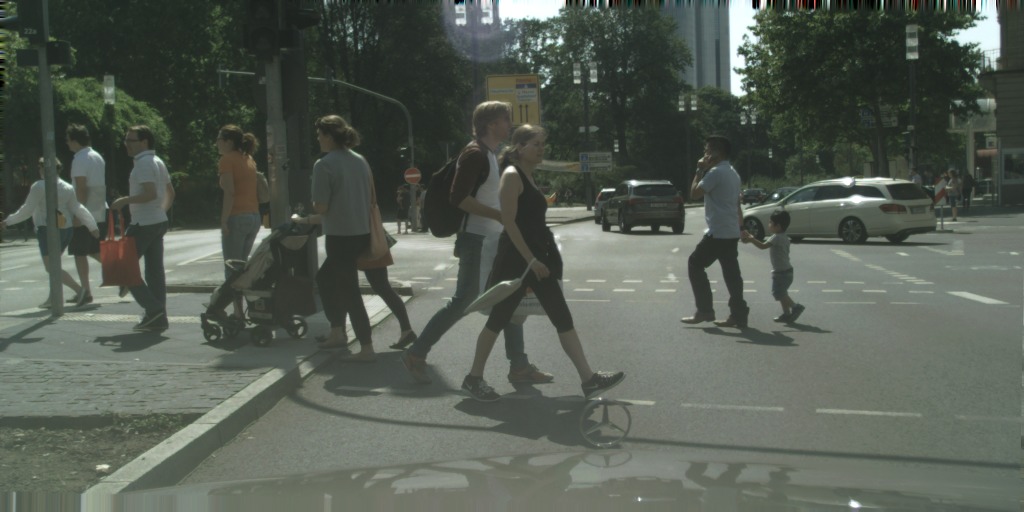} & 
\includegraphics[width=.23\linewidth]{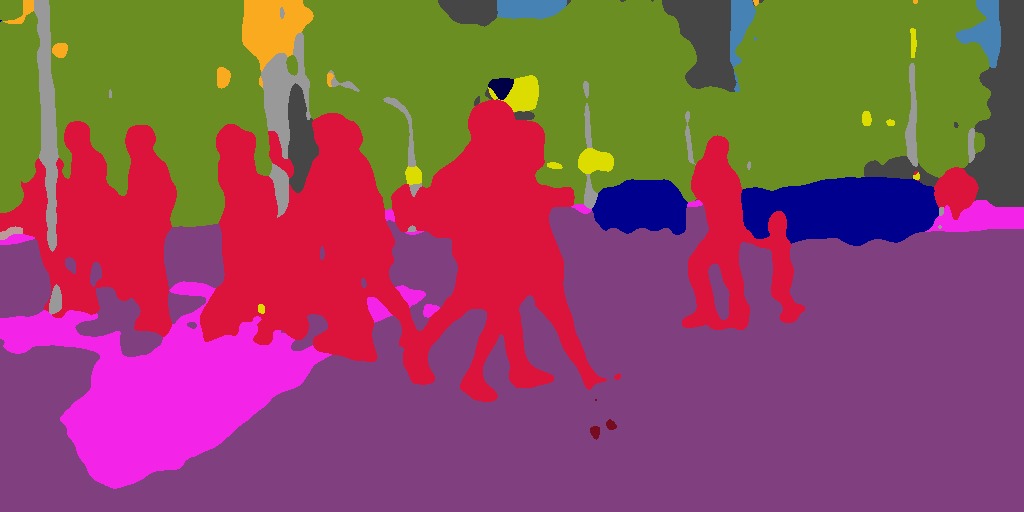} \\

(e) universal noise (4x) & (f) static adv. target & (g) universal noise (4x) & (h) static adv. target \\
\includegraphics[width=.23\linewidth]{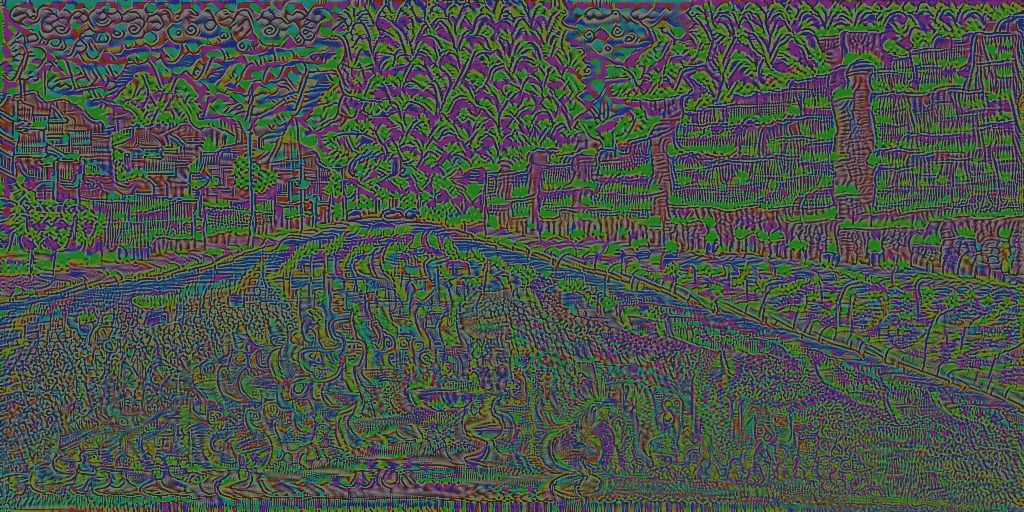} &
\includegraphics[width=.23\linewidth]{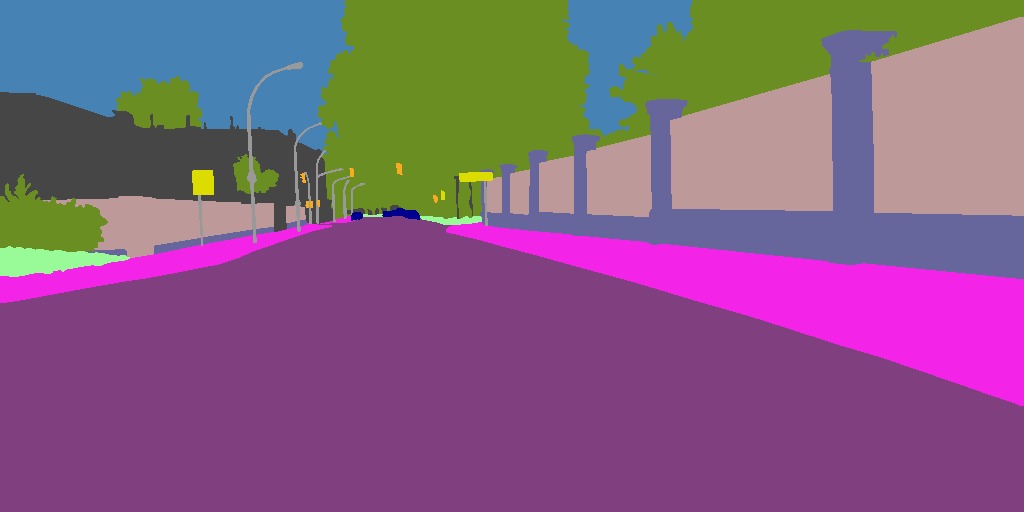} & 
\includegraphics[width=.23\linewidth]{pics/static_target/noise_eps_10_iter_60_4x} &
\includegraphics[width=.23\linewidth]{pics/static_target/adv_desired_target} \\

(i) adv. example 1 & (j) pred. adv. 1 & (k)  adv. example 2 & (l) pred. adv. 2 \\
\includegraphics[width=.23\linewidth]{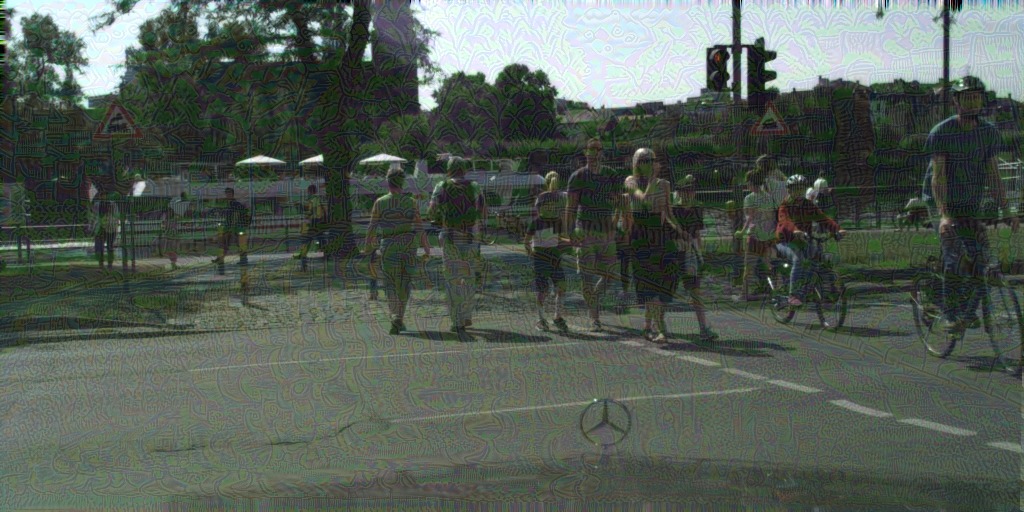} & 
\includegraphics[width=.23\linewidth]{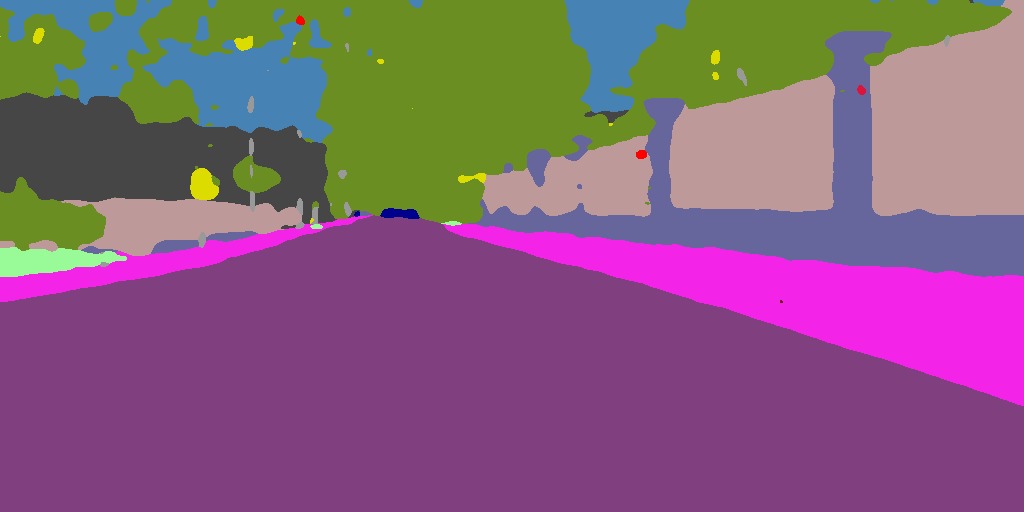} & 
\includegraphics[width=.23\linewidth]{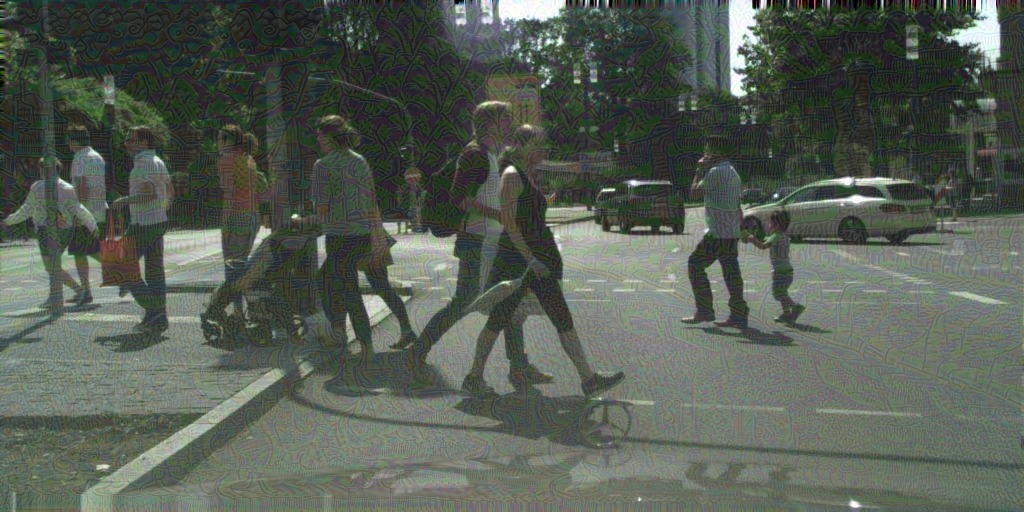} &
\includegraphics[width=.23\linewidth]{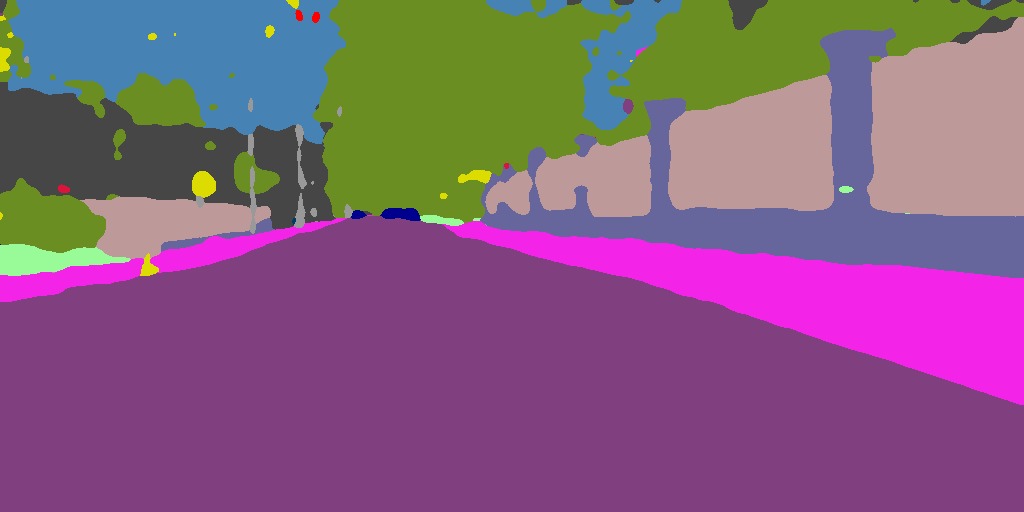}
\\
\end{tabular}
\end{center}
\caption{\emph{Influence of universal adversarial perturbation for static targets ($\varepsilon=10)$}: \textbf{(a)} First unmodified Cityscapes image. \textbf{(b)} Network prediction on (a)  \textbf{(c)} Second unmodified Cityscapes image. \textbf{(d)} Network prediction on (c) \textbf{(e)} Universal adversarial perturbation (amplified by factor 4). \textbf{(f)} Static adversarial target. \textbf{(g)} Universal adversarial perturbation (same as (e)). \textbf{(h)}  Static adversarial target (same as (f)). \textbf{(i)} Adversarial example for (a). \textbf{(j)} Network prediction on (i) \textbf{(k)} Adversarial example for (c).  \textbf{(l)} Network prediction on (k).
Please refer to the supplementary material (Section \ref{Suppl:StaticBest}-\ref{Suppl:StaticRandom}) for additional and higher-resolution illustrations.}
\label{fig:illustration_static_targets}
\end{figure*}

\paragraph{Static Target Segmentation}
As Cityscapes does not involve static scenes, we evaluated an even more challenging scenario: namely to output a static target scene segmentation which has nothing in common with the actual input scene present in the image. For this, we selected an arbitrary ground-truth segmentation (monchengladbach\_000000\_026602\_gtFine) from Cityscapes as target. We set the number of training images to $m=2975$, which corresponds to the number of images in the Cityscapes train set. Moreover, we used the unweighted loss $J_{ss}$, and did not use periodic tiles, i.e., $h=512, w=1024$. An illustration for this setting on unseen validation images is shown in Figure \ref{fig:illustration_static_targets}. The adversary achieved the desired target segmentation nearly perfectly when adding the universal perturbation that was generated on the training images. This is even more striking as for a human, the original scene, which has nothing in common with the target scene, remains clearly dominant.

\begin{figure*}
\begin{center}
\includegraphics[width=.8\linewidth]{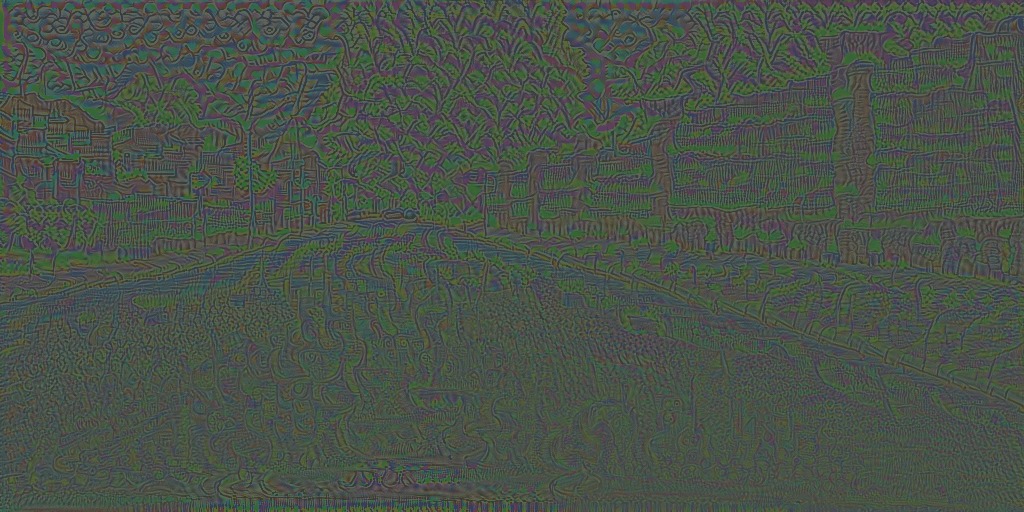}
\end{center}
\caption{Illustration of universal perturbation for a static target segmentation ($\varepsilon=20$, not amplified). Best seen in color. The network's prediction when applied to the perturbation itself as input strongly resembles the static target segmentation (see Section \ref{Suppl:PredOnPerturbation}).}
\label{fig:universal_static_perturbation}
\end{figure*}

Figure \ref{fig:universal_static_perturbation} shows an illustration of the generated universal perturbation for $\varepsilon=20$. This perturbation is highly structured and the local structure depends strongly on the target class. When comparing the perturbation with the static target segmentation, it is fairly easy to recognize the structure of the target in the perturbation. For instance, man-made structures such as buildings and fences correspond to mostly horizontal and vertical edges. This property indicates that the adversarial attack might exploit the (generally desirable) robustness of deep networks to contrast changes. This allows low contrast noise structures to have stronger impact than the high-contrast structures in the actual image.  

Table \ref{fig:static_target_epsilon} shows a quantitative analysis of the success rate for different values of $\varepsilon$. Here, we define the success rate as the categorical accuracy between static target segmentation and predicted segmentation of the network on the adversarial example. The success rate on training and validation data is nearly on par, which shows that overfitting is not an issue even for high-dimensional perturbations. This is probably due to the large number of training images and the consistent target. Unsurprisingly, larger $\varepsilon$ leads to higher success rates. The value $\varepsilon=10$ strikes a good balance between high success rate and being quasi-imperceptible.

\begin{table}
\begin{center}
\begin{tabular}{lcccc}
 & 2 & 5 & 10 & 20 \\
 \hline
Training data & 60.9\% & 82.0\% & 92.7\% & 97.2\% \\
Validation data & 60.9\% & 80.3\% & 91.0\% & 96.3\%
\end{tabular}
\end{center}
\caption{Success rate of static target segmentation for different values of $\varepsilon$. The generated perturbations achieve nearly the same success rate on unseen validation data as on the training data.}
\label{fig:static_target_epsilon}
\end{table}

\begin{figure*}
\begin{center}
\setlength{\tabcolsep}{3pt}
\begin{tabular}{cccc}
(a) image 1 & (b) pred. image 1 & (c) image 2 & (d) pred. image 2 \\
\includegraphics[width=.23\linewidth]{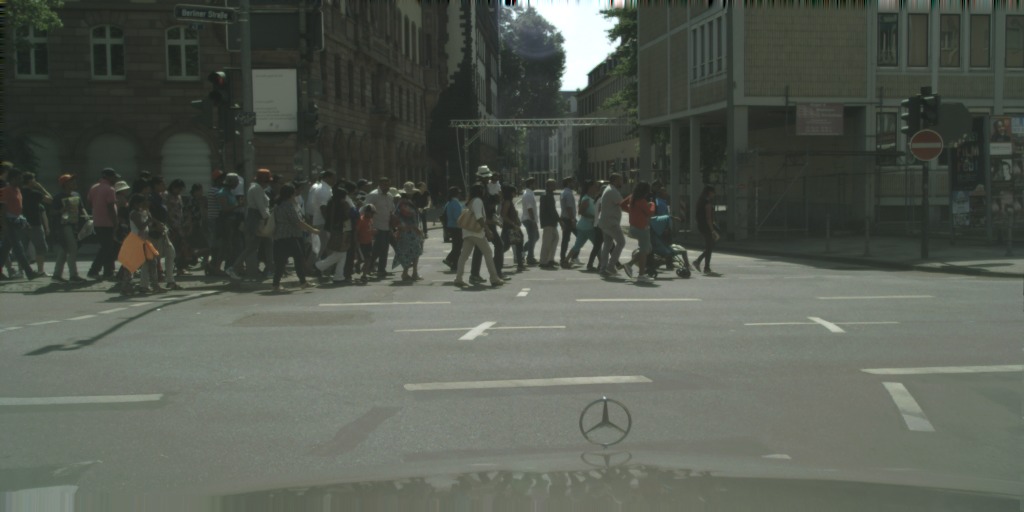} & 
\includegraphics[width=.23\linewidth]{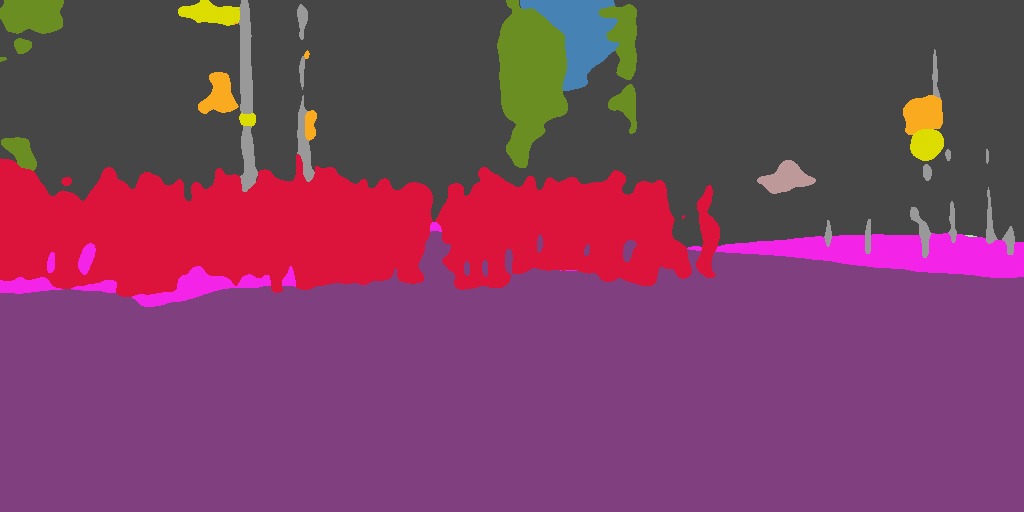} & 
\includegraphics[width=.23\linewidth]{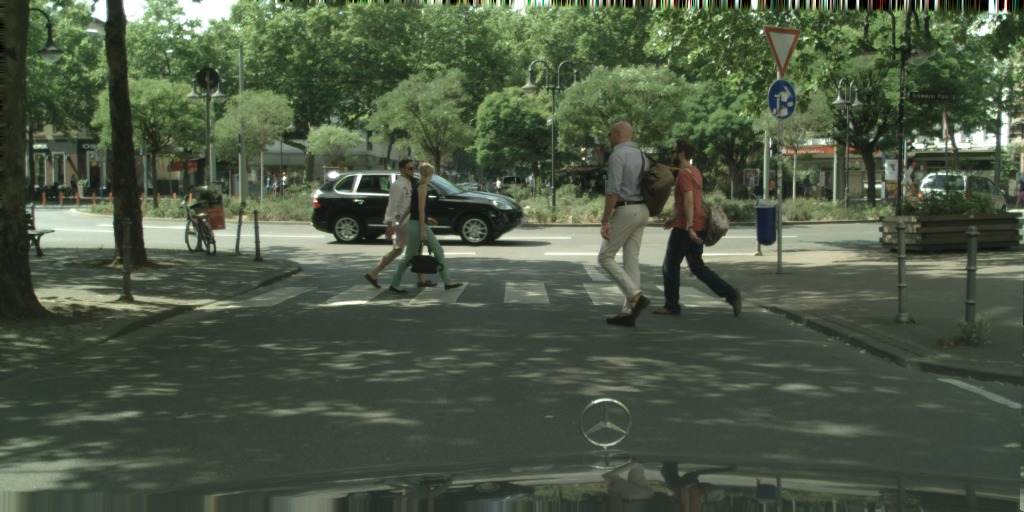} & 
\includegraphics[width=.23\linewidth]{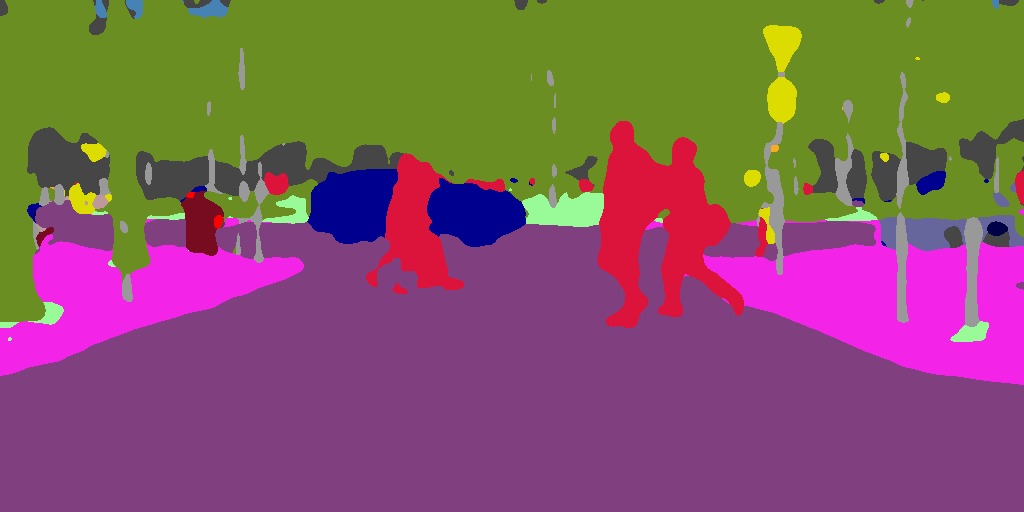} \\
 \\

(e) universal noise (4x)  & (f) dynamic adv. target 1 & (g) universal noise (4x) & (h) dynamic adv. target 2 \\
\includegraphics[width=.23\linewidth]{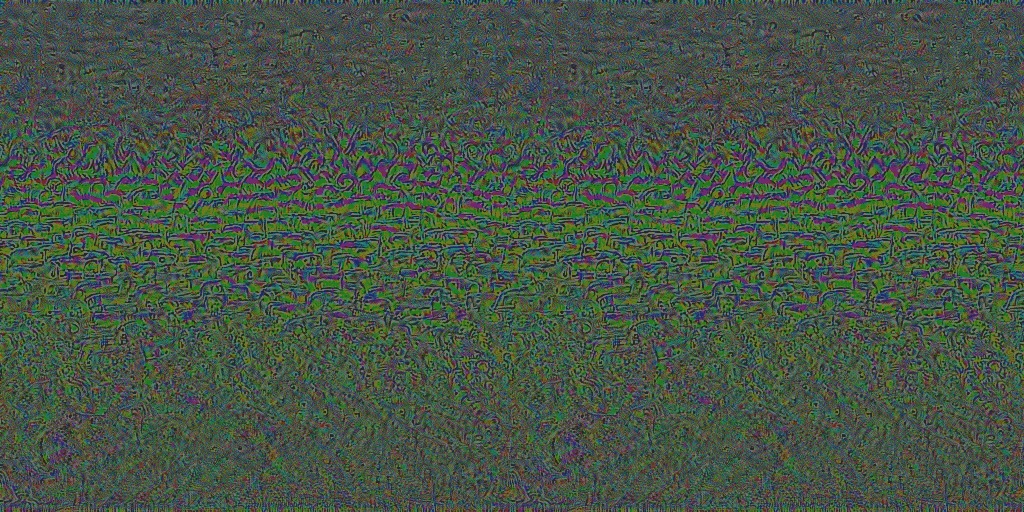} &
\includegraphics[width=.23\linewidth]{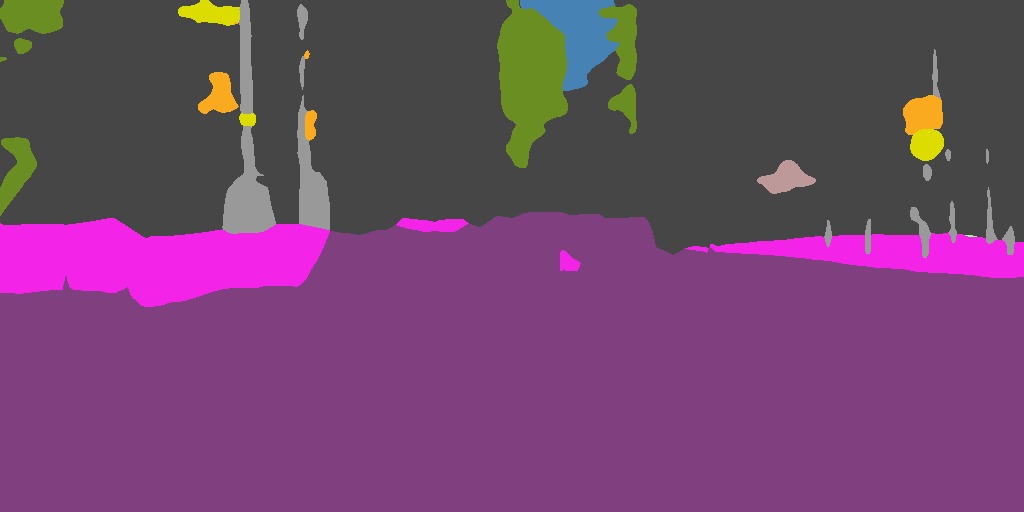} &
\includegraphics[width=.23\linewidth]{pics/dynamic_target/noise_eps_10_iter_60_4x} &
\includegraphics[width=.23\linewidth]{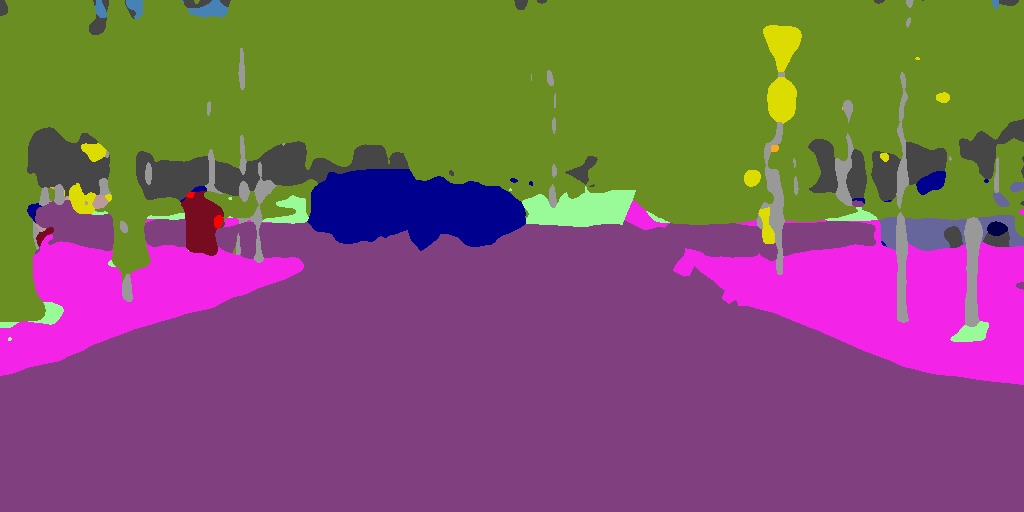} \\

(i) adv. example 1 & (j) pred. adv. 1 & (k) adv. example 2 & (l) pred. adv. 2 \\
\includegraphics[width=.23\linewidth]{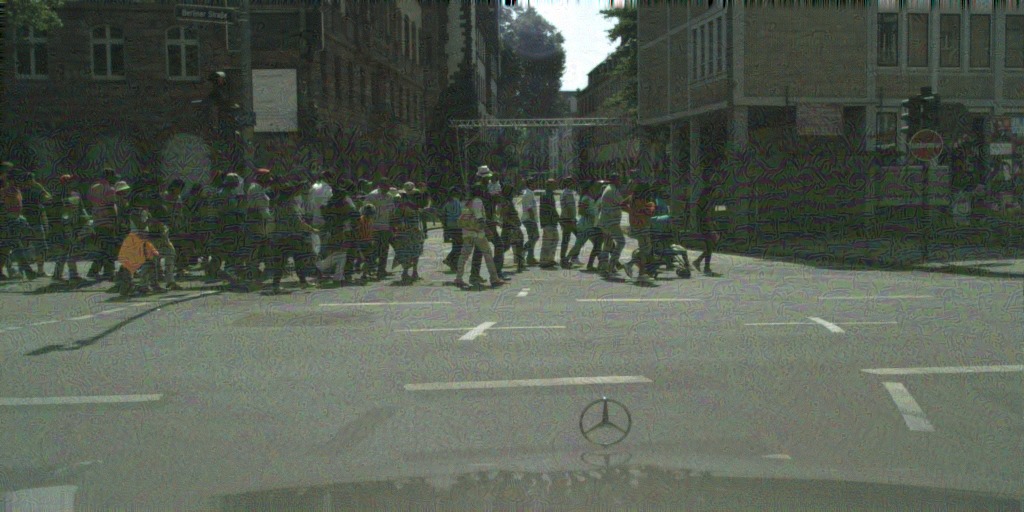} & 
\includegraphics[width=.23\linewidth]{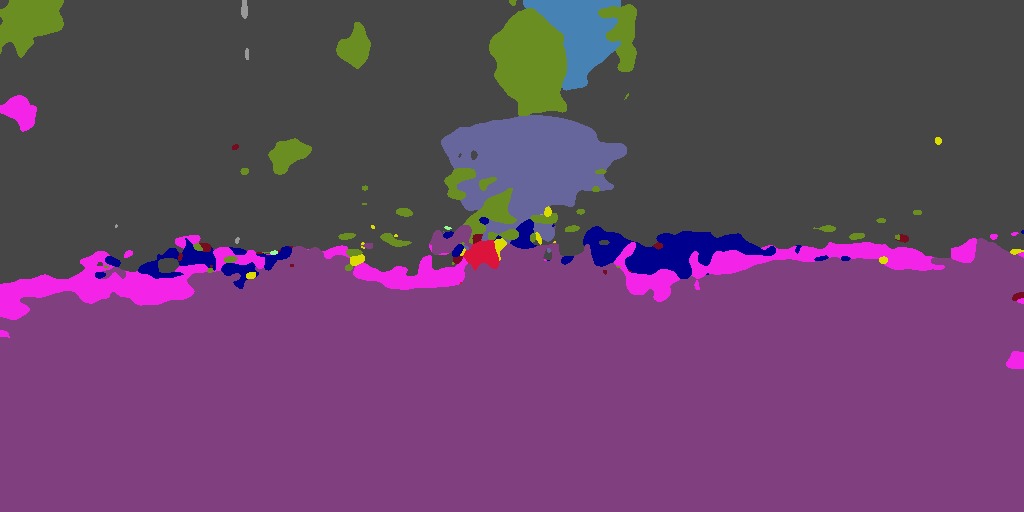} & 
\includegraphics[width=.23\linewidth]{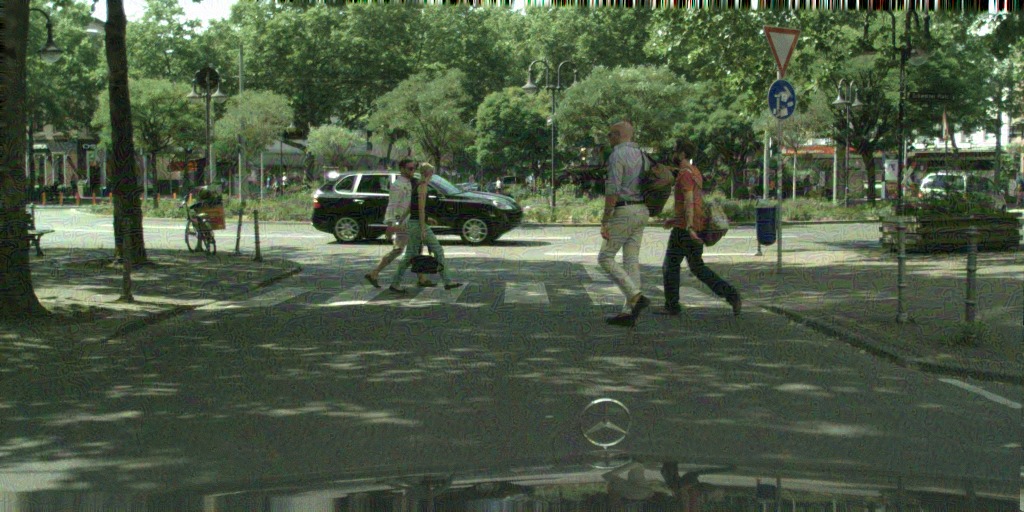} &
\includegraphics[width=.23\linewidth]{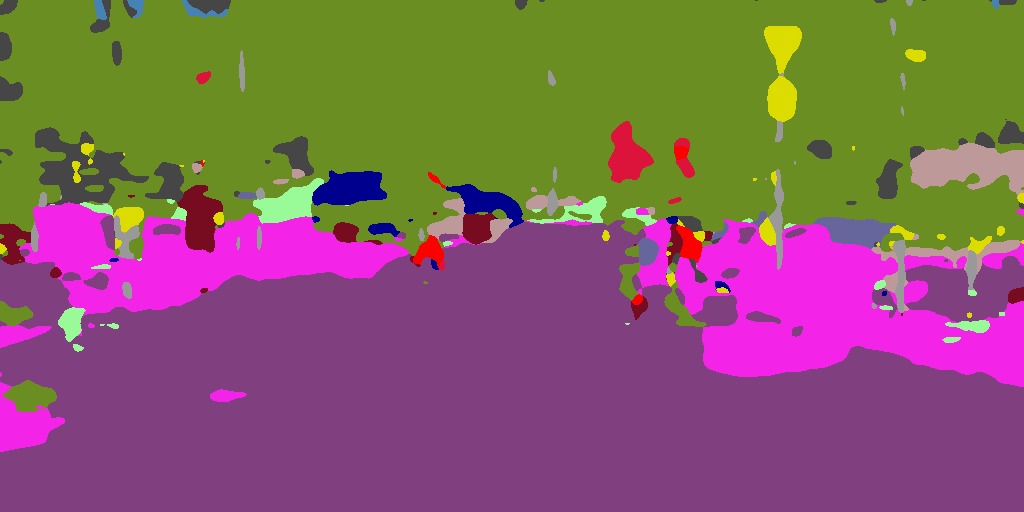} \\
\end{tabular}
\end{center}
\caption{\emph{Influence of universal adversarial perturbation for dynamic targets ($\varepsilon=10$)}: \textbf{(a)} First unmodified Cityscapes image. \textbf{(b)} Network prediction on (a). \textbf{(c)} Second unmodified Cityscapes image. \textbf{(d)} Network prediction on (c). \textbf{(e)} Universal adversarial perturbation (amplified by factor 4). \textbf{(f)} Dynamic adversarial target for (a). Note that the adversary does not tailor the universal perturbation to this target for validation data; the image solely shows the ideal output.  \textbf{(g)} Universal adversarial perturbation (same as (e)). \textbf{(h)} Dynamic adversarial target for (c). \textbf{(i)} Adversarial example for (a). \textbf{(j)} Network prediction on (i). \textbf{(k)} Adversarial example for (c). \textbf{(l)} Network prediction on (k). Please refer to the supplementary material (Section \ref{Suppl:DynamicBest}-\ref{Suppl:DynamicRandom}) for additional and higher-resolution illustrations.
}
\label{fig:illustration_dynamic_targets}
\end{figure*}

\begin{figure}
\centering
\includegraphics[width=\linewidth]{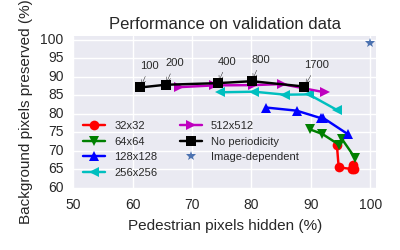}
\caption{Evaluation of universal perturbations on dynamic target segmentation for different tile-sizes and number of train images (between 100 and 1700) on validation data ($\varepsilon=10$, $\omega=0.9999$). More training images improve generalization to validation data. Smaller tile sizes increase the percentage of pedestrian pixels removed at the cost of preserving the background less well. For comparison, image-dependent non-periodic perturbations are also shown, which nearly perfectly achieve both objectives. }
\label{fig:dynamic_target_evaluation}
\end{figure}

\paragraph{Dynamic Target Segmentation}
In this experiment, we focused on an adversary which tries to hide all pedestrians (Cityscapes class ``person'') in an image while leaving the segmentation unchanged otherwise. When not noted otherwise, we set the number of training images to $m=1700$ (this value corresponds to the number of images containing pedestrians in the Cityscapes train set), the periodic tile size to $h=w=512$ and use $J^\omega_{ss}$ with $\omega=0.9999$ as motivated empirically (see Figure \ref{fig:dynamic_target_evaluation} and Table \ref{fig:dynamic_target_epsilon} and \ref{fig:dynamic_target_omega}). An illustration for this setting on unseen validation images is shown in Figure \ref{fig:illustration_dynamic_targets}. We note that qualitatively, the adversary succeeds in removing nearly all pedestrian pixels while leaving the background mostly unchanged. However, closer inspection by a human would probably raise suspicion as the predicted segmentation looks relatively inhomogeneous.

For quantifying how well an adversary achieves its primary objective of hiding a target class, we measure which percentage of the pixels that were predicted as pedestrians on the original input are assigned to any of the other classes for the adversarial example (``Pedestrian pixels hidden''). We measure the categorical accuracy on background pixels (i.e., pixels that were not predicted as pedestrians on the original input) between dynamic adversarial target segmentation and the segmentation predicted by the network on the adversarial example (``Background pixels preserved''). This quantifies the secondary objective of being inconspicuous by preserving the background. Note that this comparison does not involve the ground-truth segmentation; we solely measure if the network's original background segmentation is preserved.

\begin{table}
\begin{center}
\begin{tabular}{lcccc}
 & 2 & 5 & 10 & 20 \\
 \hline
Pedestrian hidden & 40\% & 93\% & 100\% & 100\% \\
Background pres. & 95\% & 84\% & 87\% & 89\% \\
\hline
Pedestrian hidden & 34\% & 81\% & 92\% & 93\% \\
Background pres. & 94\% & 85\% & 86\% & 87\%
\end{tabular}
\end{center}
\caption{Dynamic target for different values of $\varepsilon$ on training data (top) and validation data (bottom).}
\label{fig:dynamic_target_epsilon}
\end{table}

\begin{table}
\begin{center}
\begin{tabular}{lccccc}
 & no & 0.9 & 0.99 & 0.999 & 0.9999 \\
 \hline
Pedestrian hidden & 41\% & 70\% & 83\% & 88\% & 92\% \\
Background pres. & 96\% & 94\% & 91\% & 89\% & 86\% 
\end{tabular}
\end{center}
\caption{Dynamic target for different values of $\omega$ on validation data.}
\label{fig:dynamic_target_omega}
\end{table}

Figure \ref{fig:dynamic_target_evaluation} shows how the periodic tile-size and $m$, the number of training images, affects the results of the adversary. In general, more training images and smaller tile-sizes increase the number of hidden pedestrian pixels. This indicates that failures in hiding pedestrian pixels on validation data are mostly due to overfitting to the training data; in fact the adversary succeeds in hiding nearly $100\%$ of all pedestrian pixels on the train set for any combination of number of training images and tile-size (not shown). The number of background pixels preserved typically decreases with increased score on hiding pedestrians. As this is also the case on training images, it is likely an underfitting or optimization issue which could be improved in the future by alternative regularization methods (other than periodic noise) or more sophisticated adversarial attacks. For the presented method and $m=1700$ , a tile-size of $512\times 512$ achieves a good trade-off and is used in the remaining experiments.

Table \ref{fig:dynamic_target_epsilon} illustrates the influence of the maximum noise level $\varepsilon$. Values of $\varepsilon$ below $10$ clearly correspond to an underfitting regime as the adversary is not capable of hiding all pedestrian pixels on the train data. For $\varepsilon=10$, failures of the adversary in hiding pedestrian pixels on validation data are mostly due to overfitting (see above). Additional capacity  in the perturbation ($\varepsilon=20$) is then used by the adversary to preserve the background even better but does not help in reducing overfitting. The influence of parameter $\omega$, which allows controlling the trade-off between the primary and secondary objective, is investigated in Table \ref{fig:dynamic_target_omega}: the larger $\omega$, the more pedestrian pixels are hidden (but the background is preserved less well). Since the number of pedestrian pixels is considerably smaller than the number of background pixels, setting $\omega$ close to $1$, e.g., $\omega=0.9999$ presents a reasonable trade-off. In contrast, the unweighted loss $J_{ss}$ with no $\omega$ ($\omega=$ no) fails since it focuses too much on preserving the background.

\paragraph{Generalizability} We have tested the effect of the universal perturbation generated for Cityscapes on CamVid \cite{brostow_semantic_2009} (without any fine-tuning on CamVid). An average of 78\% of the pixels are transformed to the adversarial target for the static target segmentation. For dynamic target segmentation, an average of 84.5\% pedestrian pixels are hidden and 79.6\% of the background pixels are preserved. Thus, the perturbations generalize to a similar dataset with only a small decrease in performance. Moreover,  we have evaluated the FCN's static target universal perturbation on a PSPNet \cite{zhao_pyramid_2016}. Adding the universal perturbation reduced the IoU between PSPNet's predictions and the ground truth on Cityscapes from 75.8\% to 8.8\%. However, the IoU between the prediction and the adversarial target was also only 9.5\%. In summary, the universal perturbation generalizes over networks as an untargeted attack but not as a targeted attack.

\section{Conclusion and Outlook}
We have proposed a method for generating universal adversarial perturbations that change the semantic segmentation of images in close to arbitrary ways: an adversary can achieve (approximately) the same desired static target segmentation for arbitrary input images that have nothing in common. Moreover, an adversary can blend out certain classes (like pedestrians) almost completely while leaving the rest of the class map nearly unchanged. These results emphasize the necessity of future work to address how machine learning can become more robust against (adversarial) perturbations \cite{zheng_improving_2016,papernot_distillation_2016,kurakin_adversarial_2017} and how adversarial attacks can be detected \cite{metzen_detecting_2017,feinman_detecting_2017}. This is especially important in safety- or security-critical applications. On the other hand, the presented method does not directly allow an adversarial attack in the physical world since it requires that the adversary is able to precisely control the digital representation of the scene. While first works have shown that adversarial attacks might be extended to the physical world \cite{kurakin_adversarial_2016} and deceive face recognition systems \cite{sharif_accessorize_2016}, a practical attack against, e.g., an automated driving or surveillance system has not been presented yet. Investigating whether such practical attacks are feasible presents an important direction for future work. Furthermore, investigating whether other architectures for semantic image segmentation \cite{liu_parsing_2015, zheng_crf_rnn_2015,chen_crf_2015,fisher_dilated_2016,chen_attention_2016} are less vulnerable to adversarial perturbations is equally important.

\todo{}{Mention Tim and Fabian in acknowledgements}

{\small
\bibliographystyle{ieee}
\bibliography{references}
}

\newpage
\onecolumn
\appendix
\section{Supplementary material}
In this supplementary material, we show additional higher-resolution illustrations of Figure 3 and Figure 6. We show three examples for a static target segmentation which correspond to the images where the approach worked best (Section \ref{Suppl:StaticBest}) and worst (Section \ref{Suppl:StaticWorst}), and a randomly selected image (Section \ref{Suppl:StaticRandom}). Similarly, we show three examples for a dynamic target segmentation which correspond to the images where the approach worked best (Section \ref{Suppl:DynamicBest}) and worst (Section \ref{Suppl:DynamicWorst}), and a randomly selected image (Section \ref{Suppl:DynamicRandom}). Moreover, we also show the predictions of the network on the noise itself in Section \ref{Suppl:PredOnPerturbation}.

\section{Static Target Segmentation - Best Example} \label{Suppl:StaticBest}

\begin{center}
\setlength{\tabcolsep}{3pt}
\begin{tabular}{cc}
(a) image & (b)  prediction on image \\
\includegraphics[width=.49\linewidth]{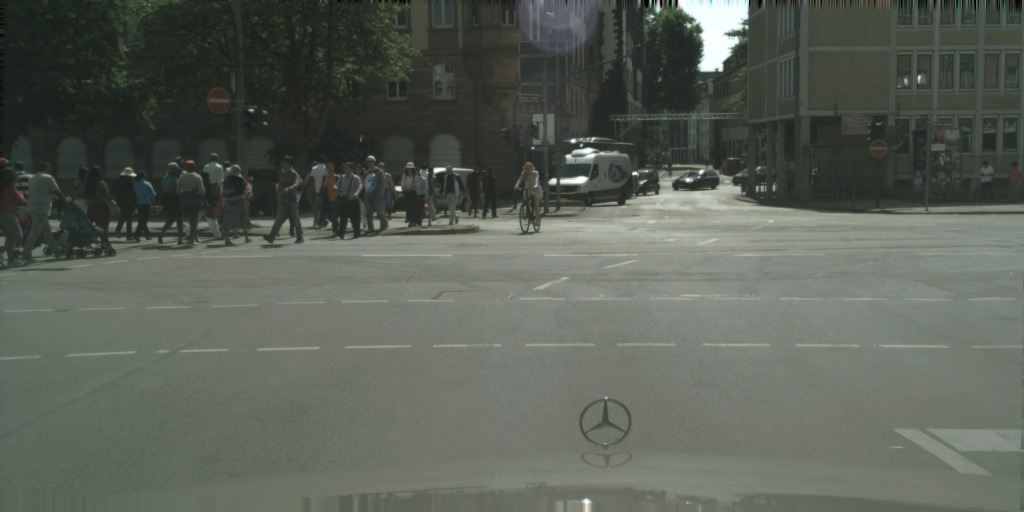} & 
\includegraphics[width=.49\linewidth]{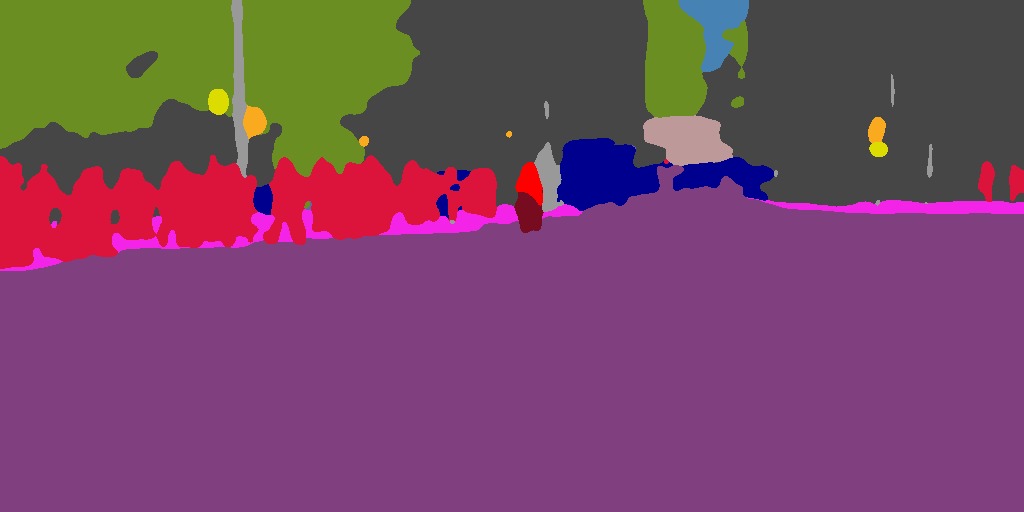} \\

(c) universal noise (4x) & (d) adv. target \\
\includegraphics[width=.49\linewidth]{pics/static_target/noise_eps_10_iter_60_4x} &
\includegraphics[width=.49\linewidth]{pics/static_target/adv_desired_target} \\

(e) adv. example & (f) pred on adv. example \\
\includegraphics[width=.49\linewidth]{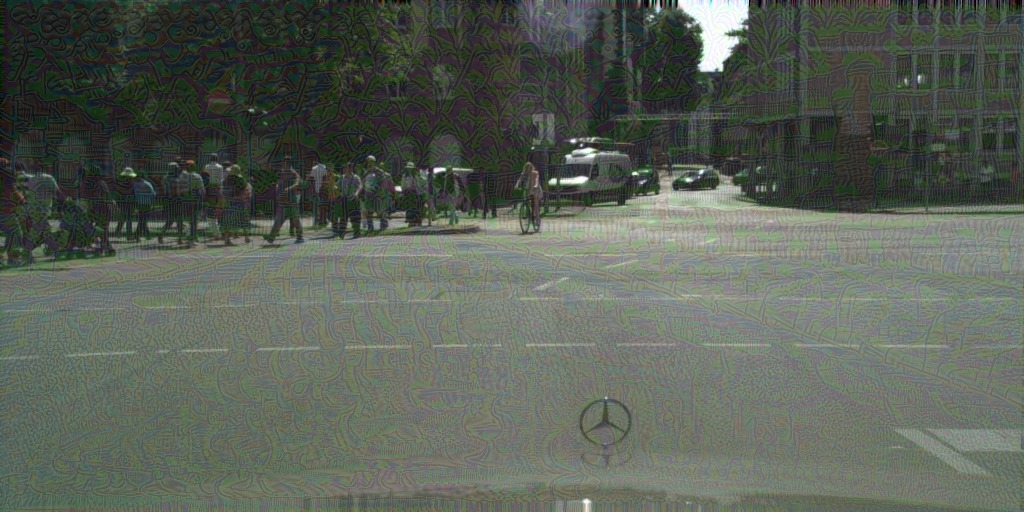} & 
\includegraphics[width=.49\linewidth]{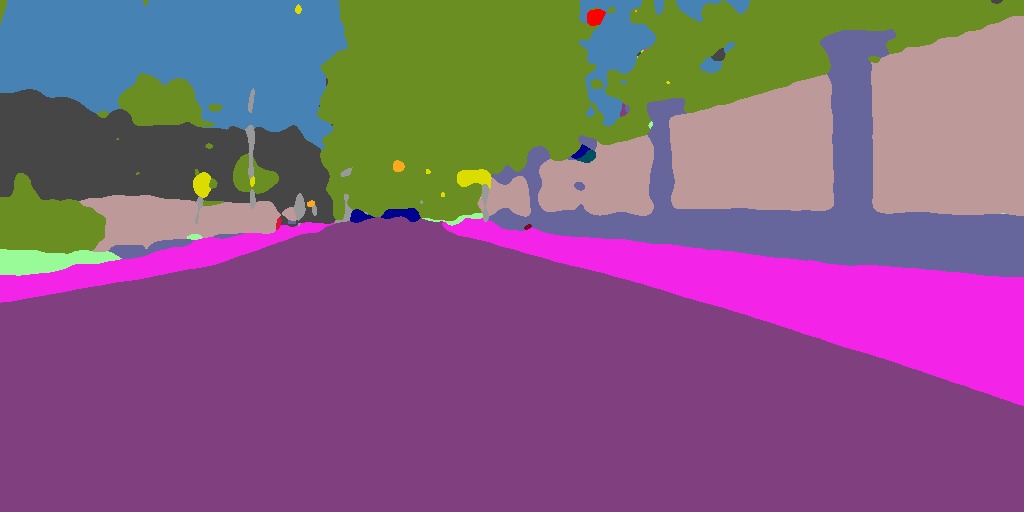} \\
\end{tabular}
\end{center}

\newpage
\section{Static Target Segmentation - Worst Example} \label{Suppl:StaticWorst}

\begin{center}
\setlength{\tabcolsep}{3pt}
\begin{tabular}{cc}
(a) image & (b)  prediction on image \\
\includegraphics[width=.49\linewidth]{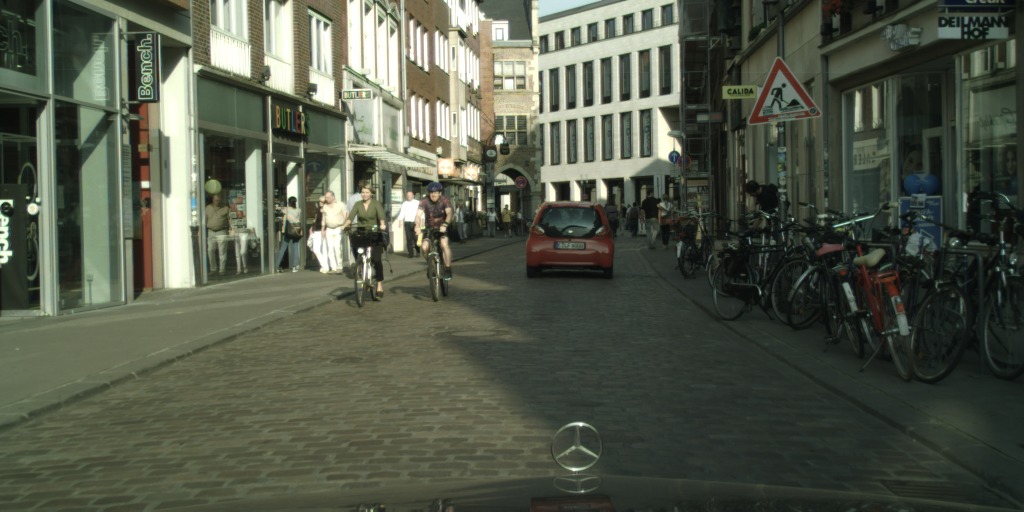} & 
\includegraphics[width=.49\linewidth]{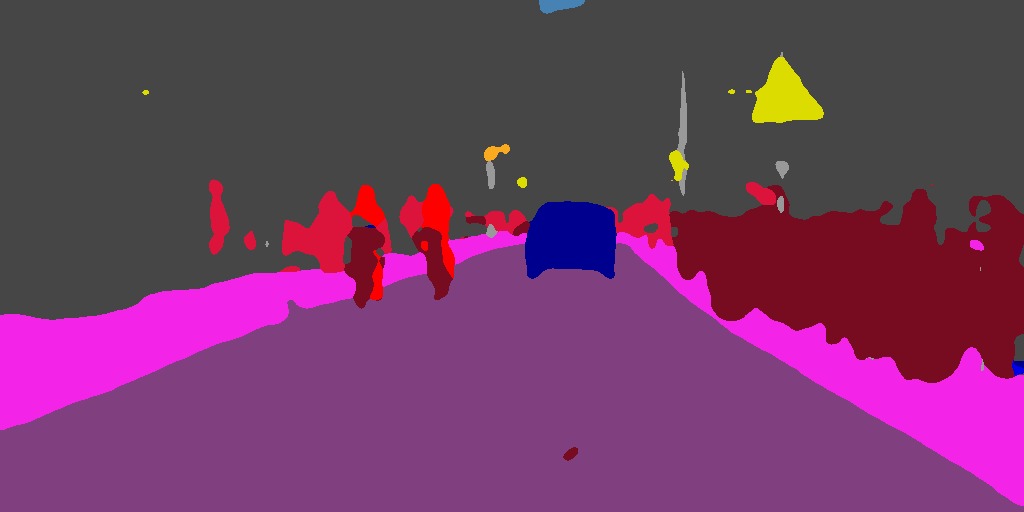} \\

(c) universal noise (4x) & (d) adv. target \\
\includegraphics[width=.49\linewidth]{pics/static_target/noise_eps_10_iter_60_4x} &
\includegraphics[width=.49\linewidth]{pics/static_target/adv_desired_target} \\

(e) adv. example & (f) pred on adv. example \\
\includegraphics[width=.49\linewidth]{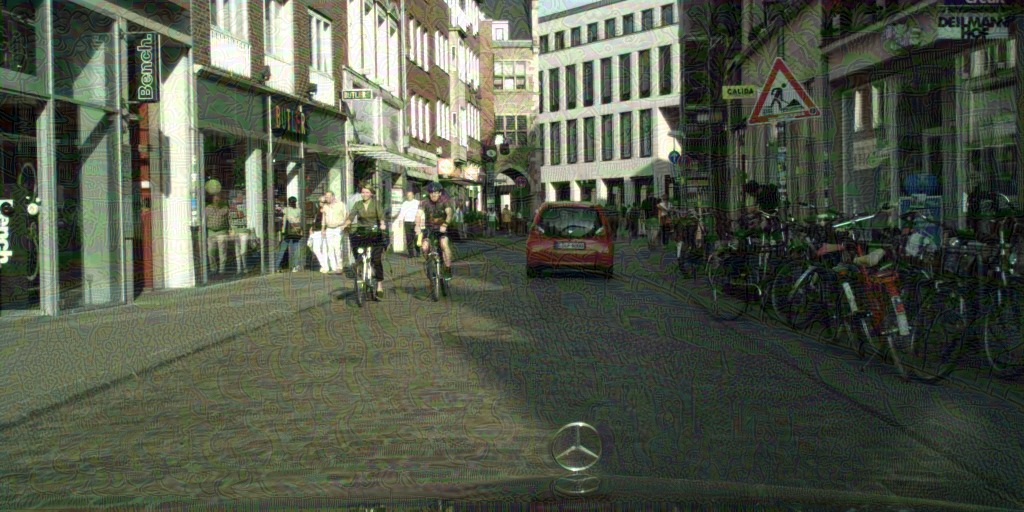} & 
\includegraphics[width=.49\linewidth]{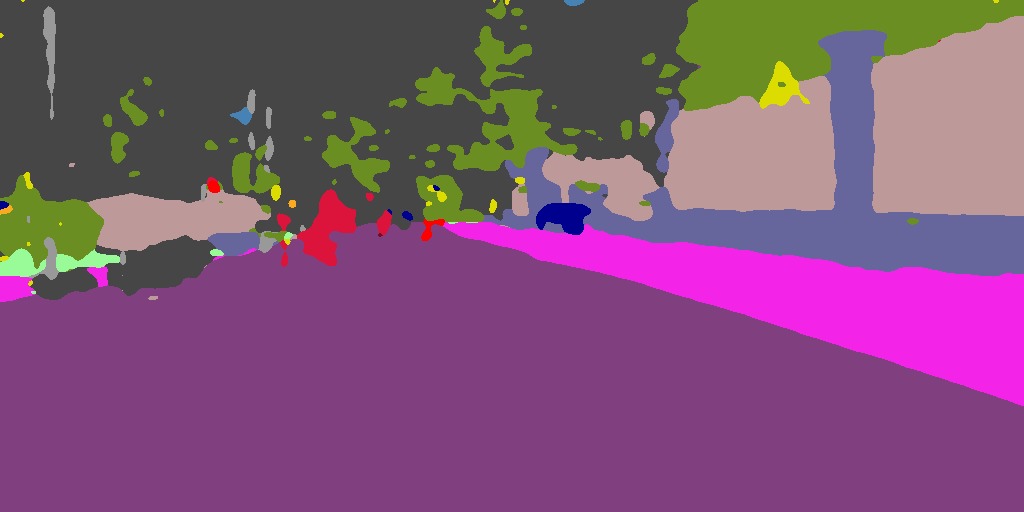} \\
\end{tabular}
\end{center}

\newpage
\section{Static Target Segmentation - Random Example} \label{Suppl:StaticRandom}

\begin{center}
\setlength{\tabcolsep}{3pt}
\begin{tabular}{cc}
(a) image & (b)  prediction on image \\
\includegraphics[width=.49\linewidth]{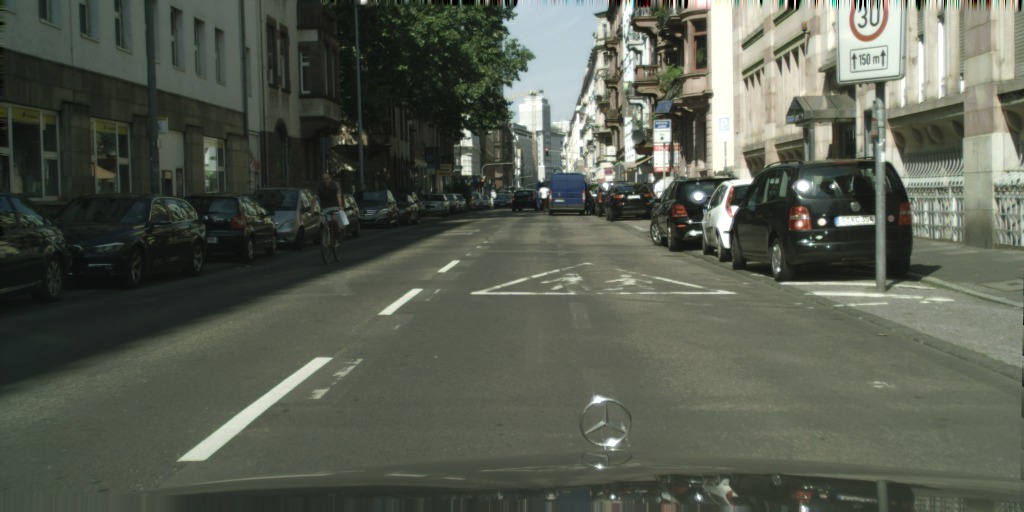} & 
\includegraphics[width=.49\linewidth]{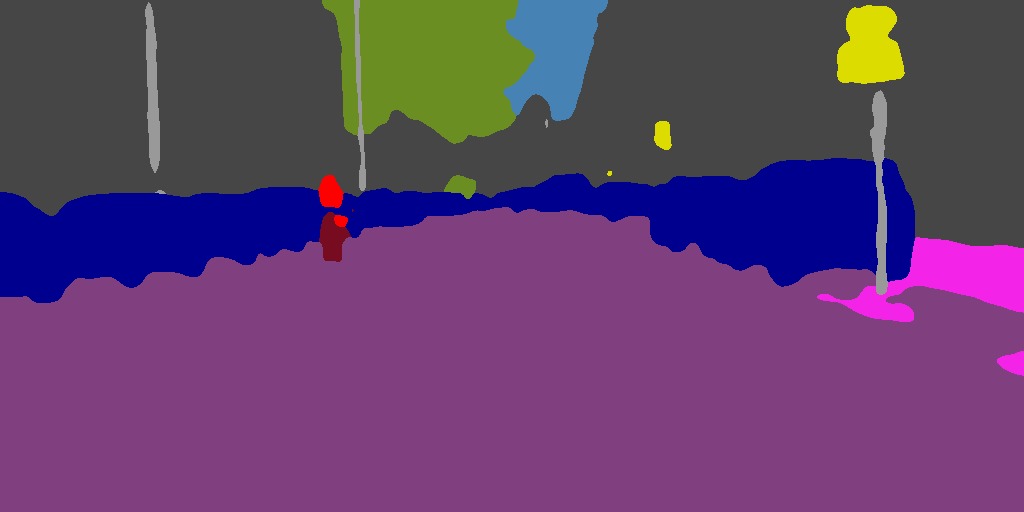} \\

(c) universal noise (4x) & (d) adv. target \\
\includegraphics[width=.49\linewidth]{pics/static_target/noise_eps_10_iter_60_4x} &
\includegraphics[width=.49\linewidth]{pics/static_target/adv_desired_target} \\

(e) adv. example & (f) pred on adv. example \\
\includegraphics[width=.49\linewidth]{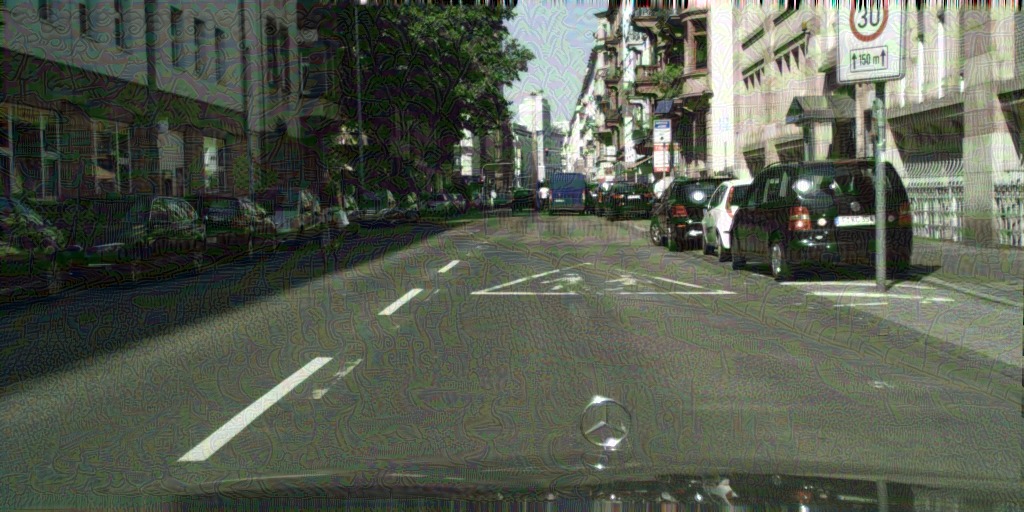} & 
\includegraphics[width=.49\linewidth]{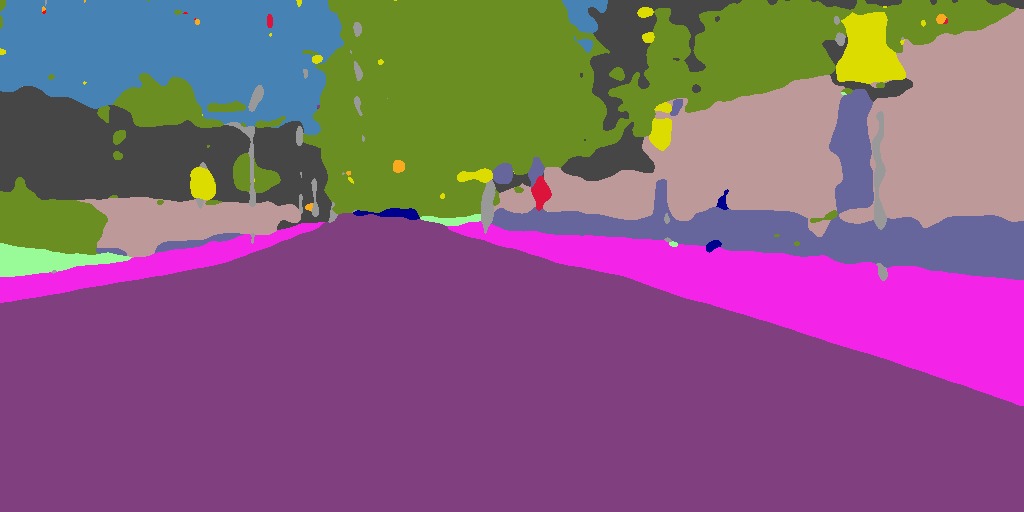} \\
\end{tabular}
\end{center}

\newpage
\section{Dynamic Target Segmentation - Best Example} \label{Suppl:DynamicBest}

\begin{center}
\setlength{\tabcolsep}{3pt}
\begin{tabular}{cc}
(a) image & (b)  prediction on image \\
\includegraphics[width=.49\linewidth]{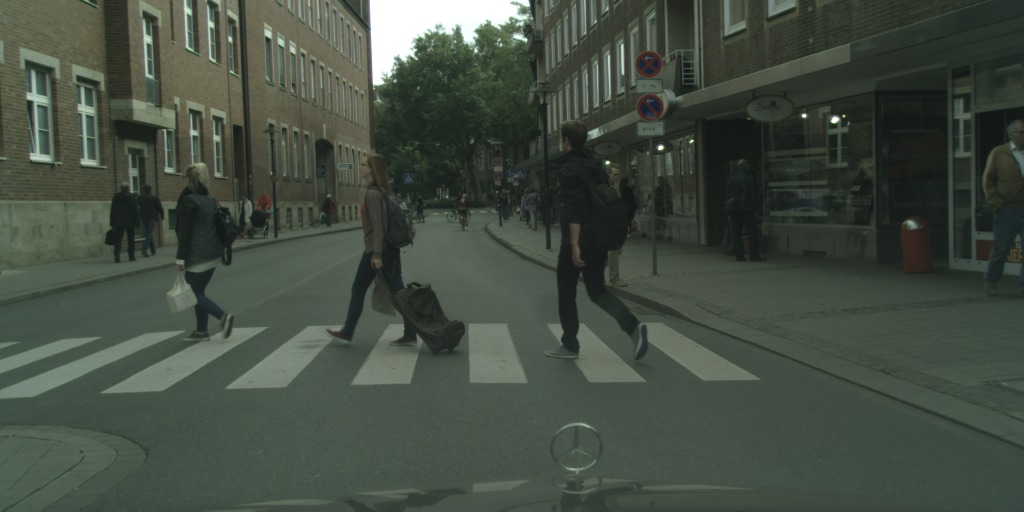} & 
\includegraphics[width=.49\linewidth]{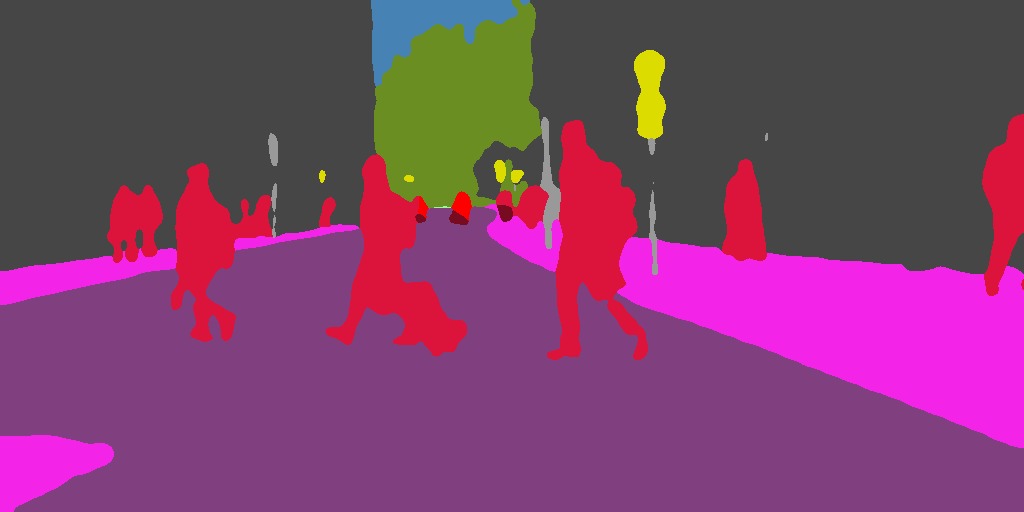} \\

(c) universal noise (4x) & (d) adv. target \\
\includegraphics[width=.49\linewidth]{pics/dynamic_target/noise_eps_10_iter_60_4x} &
\includegraphics[width=.49\linewidth]{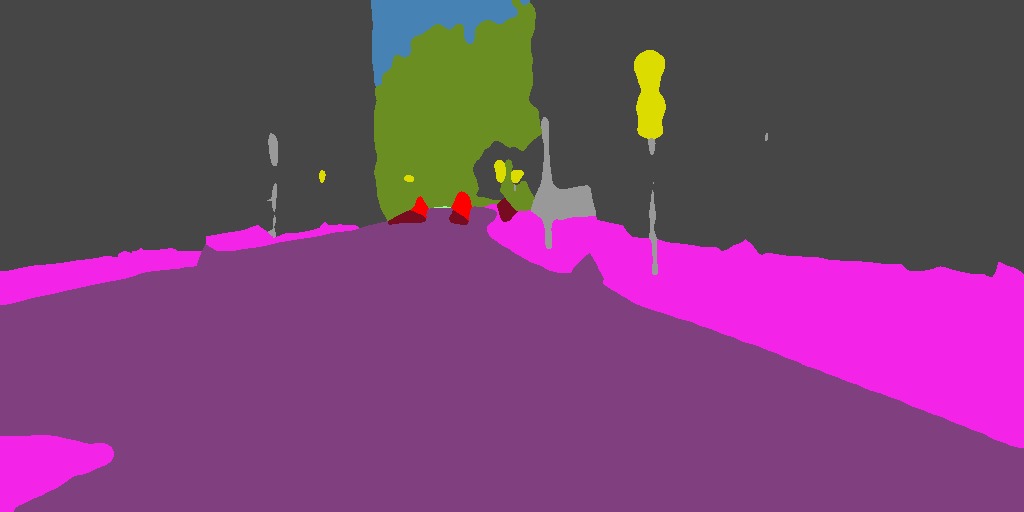} \\

(e) adv. example & (f) pred on adv. example \\
\includegraphics[width=.49\linewidth]{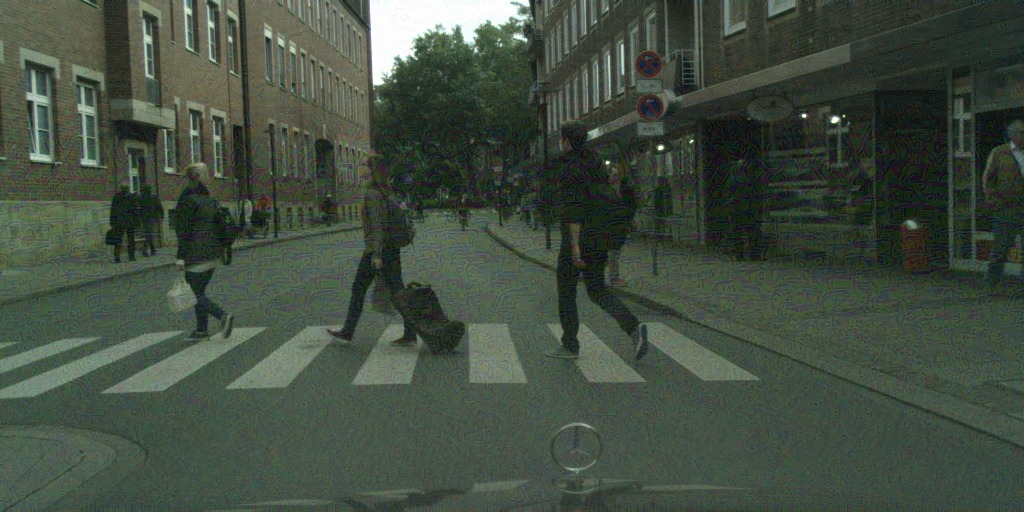} & 
\includegraphics[width=.49\linewidth]{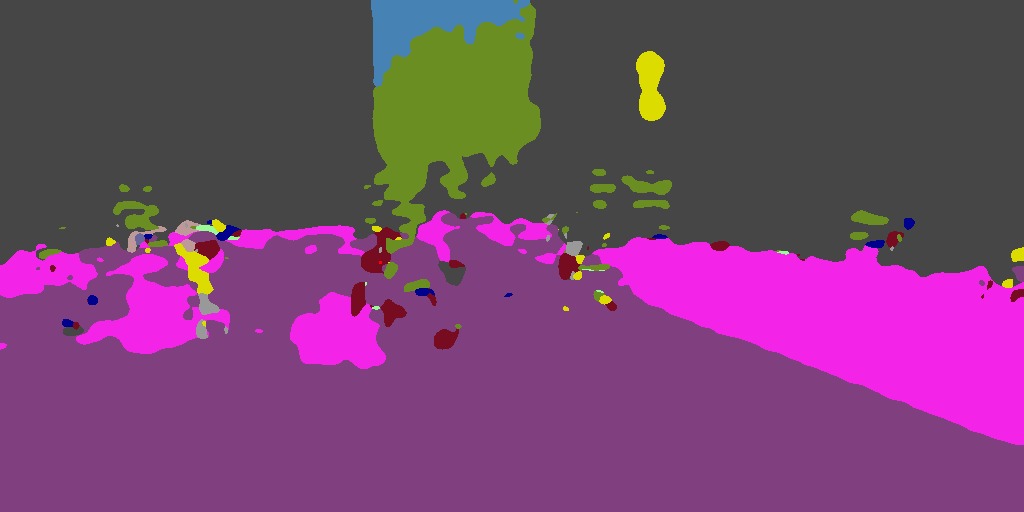} \\
\end{tabular}
\end{center}

\newpage
\section{Dynamic Target Segmentation - Worst Example} \label{Suppl:DynamicWorst}

\begin{center}
\setlength{\tabcolsep}{3pt}
\begin{tabular}{cc}
(a) image & (b)  prediction on image \\
\includegraphics[width=.49\linewidth]{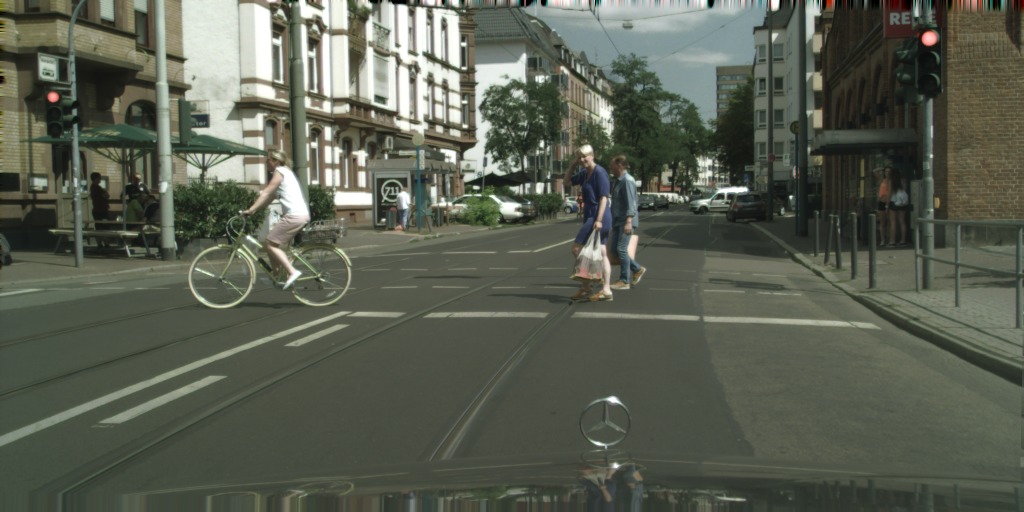} & 
\includegraphics[width=.49\linewidth]{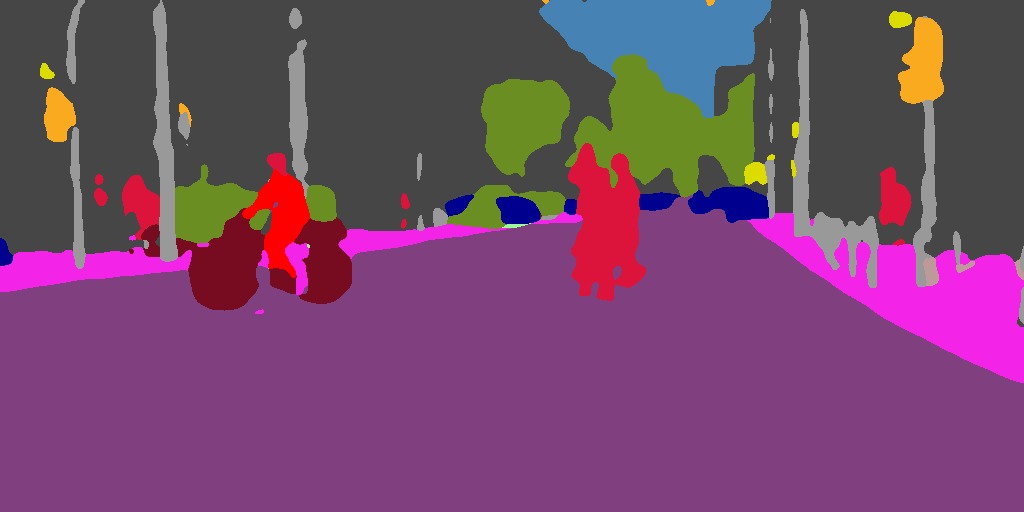} \\

(c) universal noise (4x) & (d) adv. target \\
\includegraphics[width=.49\linewidth]{pics/dynamic_target/noise_eps_10_iter_60_4x} &
\includegraphics[width=.49\linewidth]{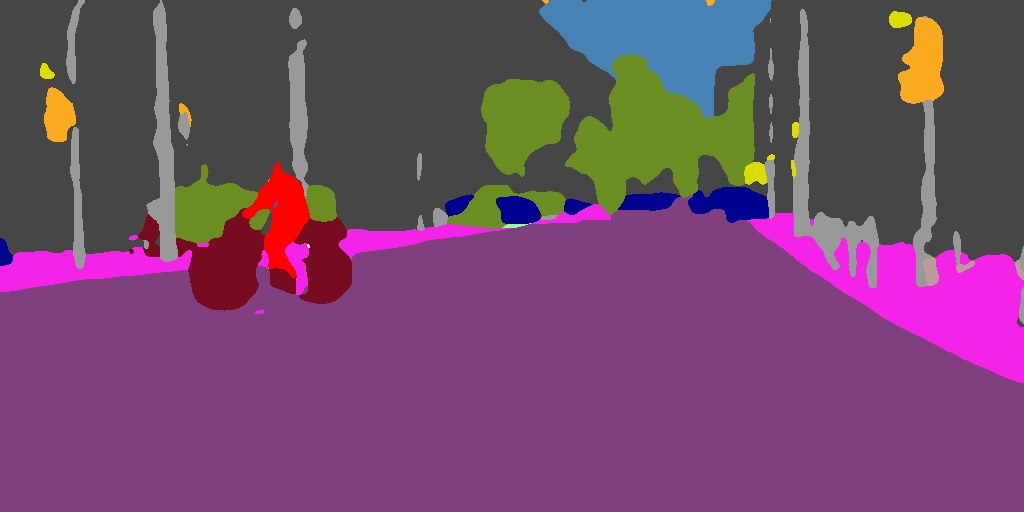} \\

(e) adv. example & (f) pred on adv. example \\
\includegraphics[width=.49\linewidth]{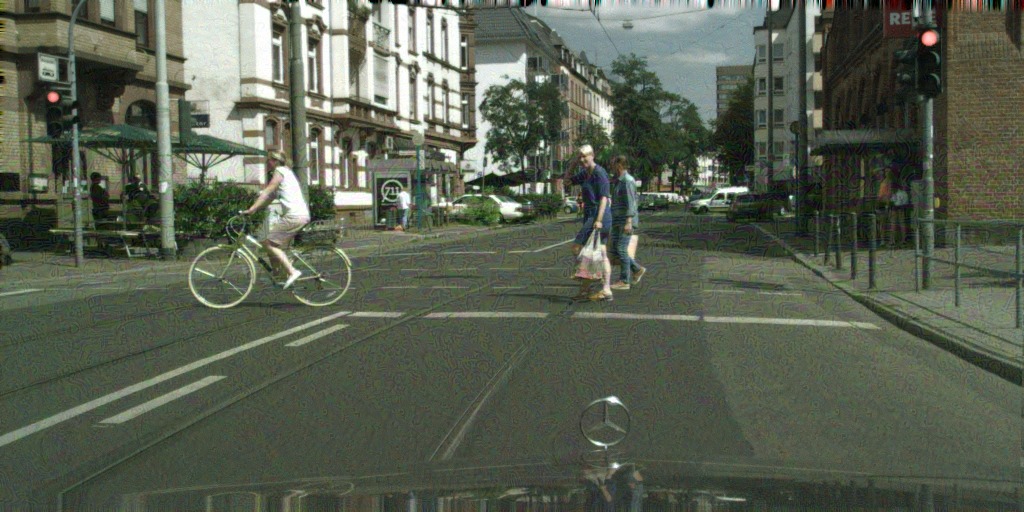} & 
\includegraphics[width=.49\linewidth]{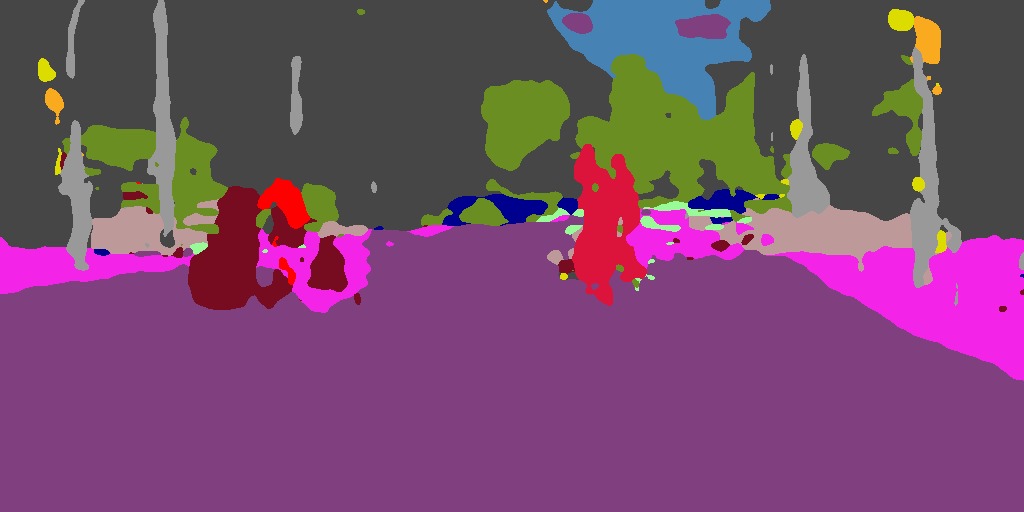} \\
\end{tabular}
\end{center}

\newpage
\section{Dynamic Target Segmentation - Random Example} \label{Suppl:DynamicRandom}

\begin{center}
\setlength{\tabcolsep}{3pt}
\begin{tabular}{cc}
(a) image & (b)  prediction on image \\
\includegraphics[width=.49\linewidth]{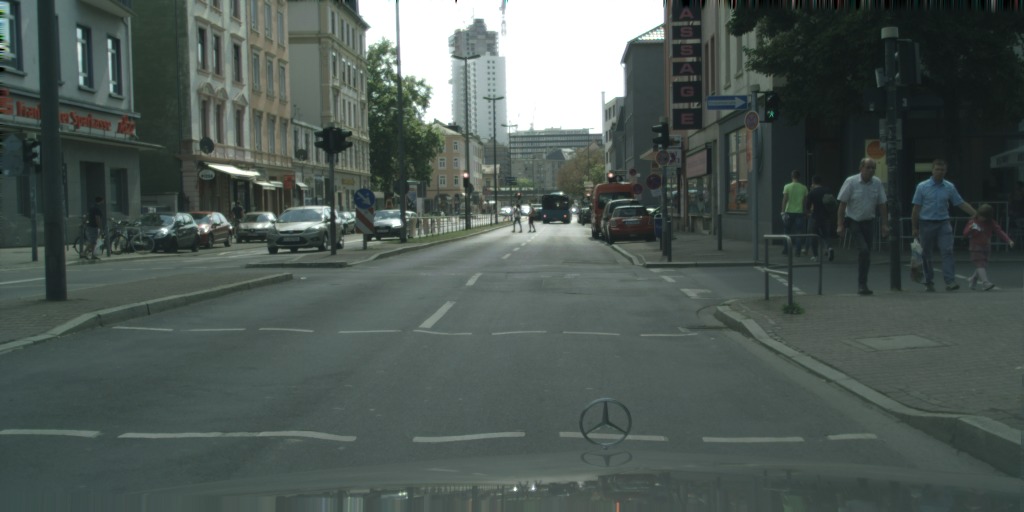} & 
\includegraphics[width=.49\linewidth]{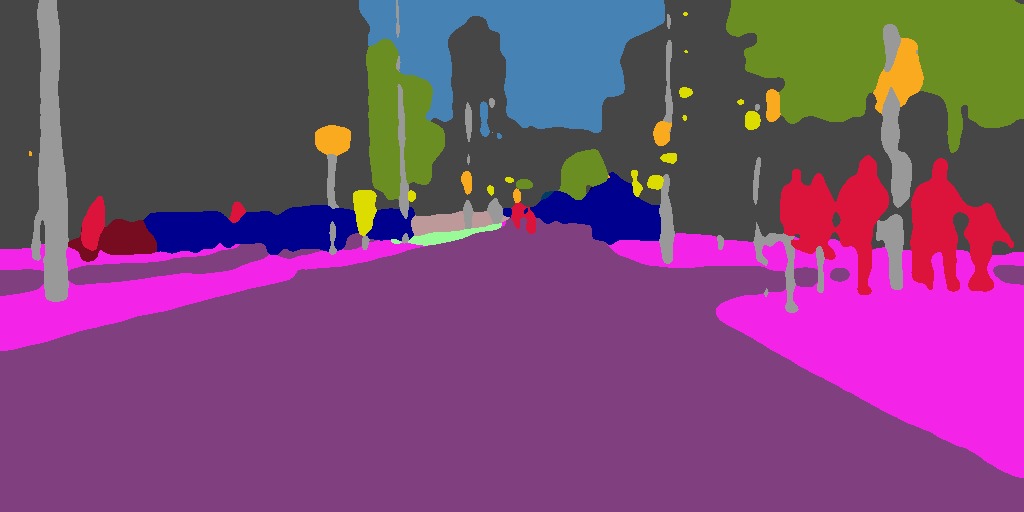} \\

(c) universal noise (4x) & (d) adv. target \\
\includegraphics[width=.49\linewidth]{pics/dynamic_target/noise_eps_10_iter_60_4x} &
\includegraphics[width=.49\linewidth]{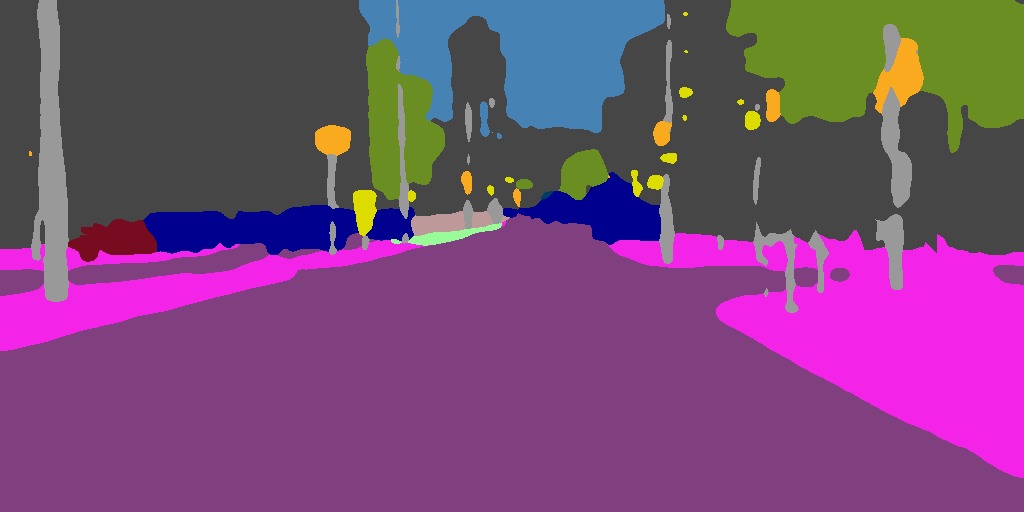} \\

(e) adv. example & (f) pred on adv. example \\
\includegraphics[width=.49\linewidth]{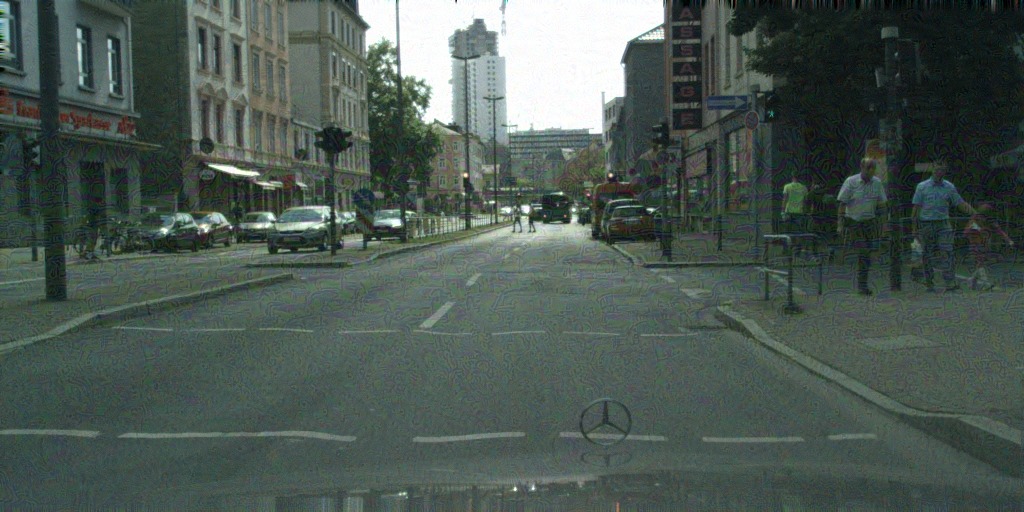} & 
\includegraphics[width=.49\linewidth]{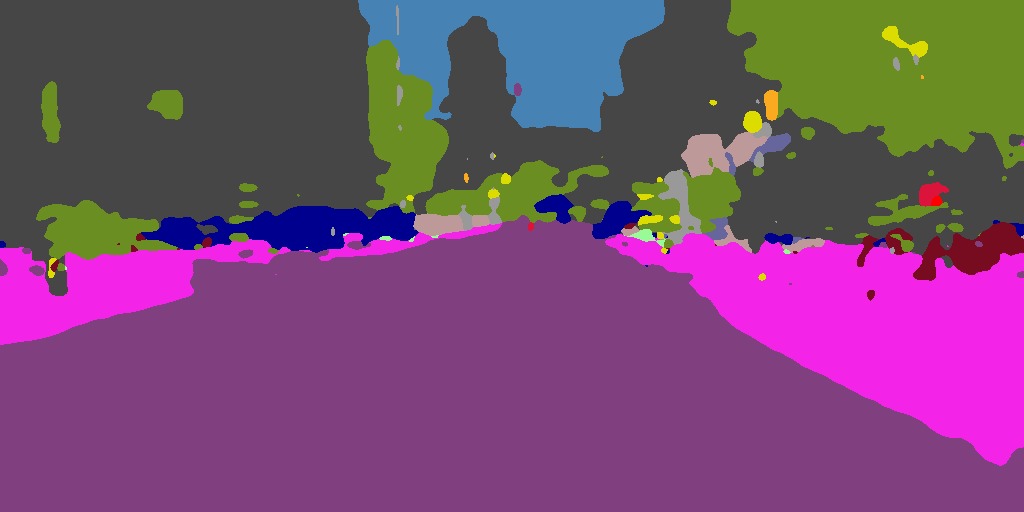} \\
\end{tabular}
\end{center}

\newpage
\section{Network Predictions on Universal Perturbations} \label{Suppl:PredOnPerturbation}

\begin{tabular}{cc}
(a) Static universal perturbation (x4) & (b) Prediction on (a) \\
\includegraphics[width=.49\linewidth]{pics/static_target/noise_eps_10_iter_60_4x} & 
\includegraphics[width=.49\linewidth]{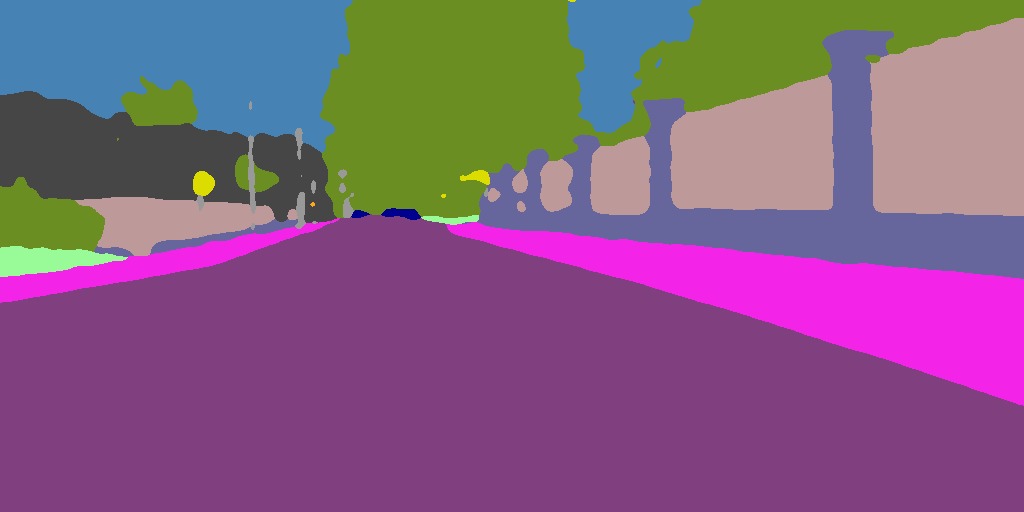} \\

(c) Dynamic universal perturbation (x4) & (d) Prediction on (c) \\
\includegraphics[width=.49\linewidth]{pics/dynamic_target/noise_eps_10_iter_60_4x} &
\includegraphics[width=.49\linewidth]{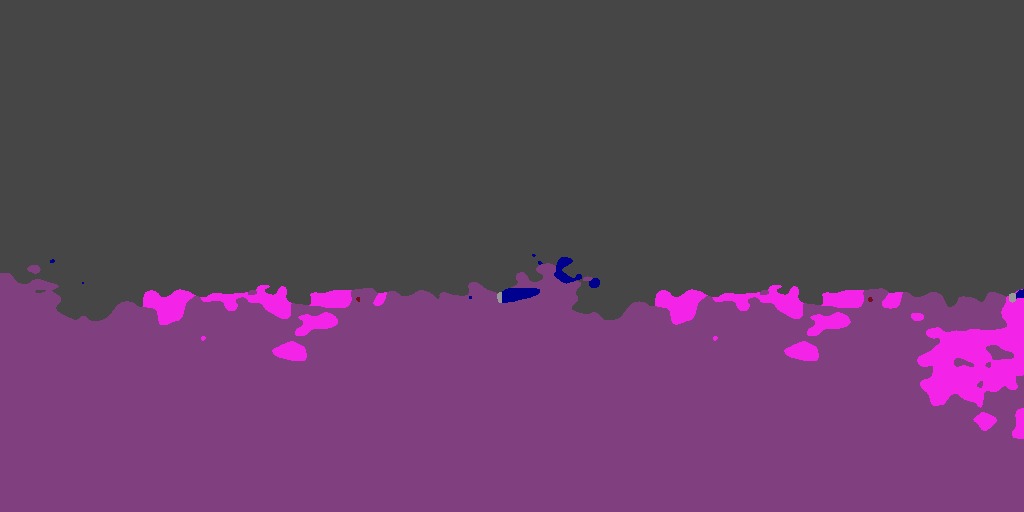} \\
\end{tabular}

\end{document}